\documentclass[10pt,journal,compsoc,x11names]{IEEEtran}
\usepackage{amsmath,amsfonts}
\usepackage{algorithm}
\usepackage{array}
\usepackage[caption=false,font=normalsize,labelfont=sf,textfont=sf]{subfig}
\usepackage{textcomp}
\usepackage{stfloats}
\usepackage{url}
\usepackage{verbatim}
\usepackage{graphicx}
\usepackage{multirow}
\usepackage{amsmath, amssymb}
\usepackage{algorithm}
\usepackage{algpseudocode}
\usepackage{ulem}
\usepackage{etoolbox}
\usepackage{amssymb}
\usepackage{xspace}
\usepackage[numbers,sort&compress]{natbib}
\usepackage{hyperref}
\hypersetup{
    colorlinks=true,      
    linkcolor=blue,       
    citecolor=blue,        
    filecolor=magenta,    
    urlcolor=blue        
}

\usepackage{forest}
\usetikzlibrary{shadows}
\usepackage{amsmath}

\usepackage{amssymb}
\usepackage{cleveref}

\definecolor{deepred}{rgb}{0.698,0.133,0.133}
\definecolor{blue}{rgb}{0,0,1}
\usepackage{tikz}
\usepackage{amssymb}

\definecolor{lightcoral}{rgb}{0.94, 0.5, 0.5}
\definecolor{lightgreen}{rgb}{0.56, 0.93, 0.56}
\definecolor{harvestgold}{rgb}{0.85, 0.57, 0.0}
\definecolor{brightlavender}{rgb}{0.75, 0.58, 0.89}
\definecolor{capri}{rgb}{0.0, 0.75, 1.0}
\definecolor{carminepink}{rgb}{0.92, 0.3, 0.26}
\definecolor{celadon}{rgb}{0.67, 0.88, 0.69}
\definecolor{darkpastelgreen}{rgb}{0.01, 0.75, 0.24}

\definecolor{deepred}{rgb}{0.698,0.133,0.133}
\definecolor{blue}{rgb}{0,0,1}
\definecolor{lightergray}{RGB}{230,230,230}
\definecolor{DarkRed}{RGB}{130,25,0}
\definecolor{PurpleRed}{RGB}{204,0,102}
\definecolor{DarkGreen}{RGB}{30,130,30}
\definecolor{DarkBlue}{RGB}{0,0,250}
\definecolor{DarkYellow}{RGB}{255,128,0}
\definecolor{light-gray}{gray}{0.95}
\definecolor{lightgreen}{RGB}{231,255,219}
\definecolor{lightred}{RGB}{252,231,234}
\definecolor{lightyellow}{RGB}{250,253,191}
\definecolor{lightpurple}{RGB}{229,204,255}
\definecolor{lightblue}{RGB}{229,246,254}
\definecolor{value-modification}{RGB}{250, 217, 86}
\definecolor{digit-expansion}{RGB}{216, 194, 104}
\definecolor{integer-decimal-fraction}{RGB}{240, 133, 51}
\definecolor{semantic-paraphrasing}{RGB}{85, 157, 63}
\definecolor{complexity-increasing}{RGB}{58, 120, 175}
\definecolor{question-transformation}{RGB}{174, 205, 225}
\definecolor{interference-injection}{RGB}{255,204,229}
\definecolor{remove-constrain}{RGB}{204,204,255}
\definecolor{myGreen}{RGB}{127,210,85}
\definecolor{myOrange}{RGB}{242,154,66}
\definecolor{myYellow}{RGB}{247,223,65}
\definecolor{myRed}{RGB}{232,80,43}
\definecolor{myViolet}{RGB}{162,57,102}
\definecolor{myBlue}{HTML}{4686f3}
\definecolor{myYellowv2}{HTML}{E6C802}
\definecolor{myOrangev2}{HTML}{ED8E55}
\definecolor{MyGreenv2}{HTML}{009B55}
\definecolor{MyRedv2}{HTML}{c22f2f}

\newcommand{\goodmetric}[1]{\textbf{#1}} 
\definecolor{lightcoral}{rgb}{0.94, 0.5, 0.5}
\definecolor{lightgreen}{rgb}{0.56, 0.93, 0.56}
\definecolor{harvestgold}{rgb}{0.85, 0.57, 0.0}
\definecolor{brightlavender}{rgb}{0.75, 0.58, 0.89}
\definecolor{capri}{rgb}{0.0, 0.75, 1.0}
\definecolor{carminepink}{rgb}{0.92, 0.3, 0.26}
\definecolor{celadon}{rgb}{0.67, 0.88, 0.69}
\definecolor{darkpastelgreen}{rgb}{0.01, 0.75, 0.24}

\definecolor{deepred}{rgb}{0.698,0.133,0.133}
\definecolor{blue}{rgb}{0,0,1}

\crefformat{section}{\S#2#1#3} 
\crefformat{subsection}{\S#2#1#3}
\crefformat{subsubsection}{\S#2#1#3}

\usepackage{amsthm}
\usepackage{amsmath}%
\usepackage{MnSymbol}%

\makeatletter
\newtheorem*{rep@theorem}{\rep@title}
\newcommand{\newreptheorem}[2]{%
\newenvironment{rep#1}[1]{%
 \def\rep@title{#2 \ref{##1}}%
 \begin{rep@theorem}}%
 {\end{rep@theorem}}}
\makeatother

\newreptheorem{theorem}{Theorem}

\newreptheorem{lemma}{Lemma}

\newreptheorem{proposition}{proposition}

\usepackage{bm}

\usepackage{booktabs}
\usepackage{xcolor,pifont}
\usepackage{bbding}
\usepackage{colortbl}

\usepackage[inline,shortlabels]{enumitem}

\usepackage{tikz,xcolor,hyperref}
\definecolor{lime}{HTML}{A6CE39}
\DeclareRobustCommand{\orcidicon}{%
    \begin{tikzpicture}
    \draw[lime, fill=lime] (0,0) 
    circle [radius=0.16] 
    node[white] {{\fontfamily{qag}\selectfont \tiny ID}};    \draw[white, fill=white] (-0.0625,0.095) 
    circle [radius=0.007];    \end{tikzpicture}
    \hspace{-2mm}}
\foreach \x in {A, ..., Z}{%
    \expandafter\xdef\csname orcid\x\endcsname{\noexpand\href{https://orcid.org/\csname orcidauthor\x\endcsname}{\noexpand\orcidicon}}
}

\definecolor{rred}{rgb}{0.75, 0.0, 0.0}

\newcommand{\llamaa}{\mbox{\textsc{LLaMA2-7B-chat}}\xspace}

\newcommand{\llamac}{\mbox{\textsc{LLaMA3-8B-instruct}}\xspace}
\newcommand{\Mistral}{\mbox{\textsc{Mistral-7B-instruct}}\xspace}

\begin{document}

\title{PromptCD: Test-Time Behavior Enhancement via Polarity-Prompt Contrastive Decoding}
\author{
Baolong Bi,
Yuyao Ge,
Shenghua Liu\textsuperscript{$\dagger$},
Yuchen He,
Siqian Tong,
Lizhe Chen,
Lingrui Mei,
Zehao Li, \\
Yiwei Wang,
Yujun Cai,
Ming-Hsuan Yang\orcidA{},~\IEEEmembership{Fellow,~IEEE}
and Xueqi Cheng\orcidB{},~\IEEEmembership{Fellow,~IEEE}

 \IEEEcompsocitemizethanks{
\IEEEcompsocthanksitem{This work is supported in part by the National
Key R\&D Program of China under Grant Nos.
2023YFA1011602, 
Beijing Natural Science Foundation No. 4262033,
and the National Natural Science 
Foundation of China under Grant Nos. U25B2076, 62441229, and 62377043.)}
\IEEEcompsocthanksitem{Baolong Bi, Yuyao Ge, Yuchen He, Lingrui Mei, Shenghua Liu, and Xueqi Cheng are with the Key Laboratory of Network Data Science and Technology, Institute of Computing Technology, Chinese Academy of Sciences, Beijing 100190, China, and also with the University of Chinese Academy of Sciences (e-mail: bibaolong23z@ict.ac.cn; geyuyao24z@ict.ac.cn; heye58478@gmail.com; meilingrui25b@ict.ac.cn; liushenghua@ict.ac.cn; cxq@ict.ac.cn).}
\IEEEcompsocthanksitem{Siqian Tong and Zehao Li are with the University of Chinese Academy of Sciences, Beijing, China (e-mail: tongsiqian01@gmail.com; lizehao23z@ict.ac.cn).}
\IEEEcompsocthanksitem{Lizhe Chen is with the Shenzhen International Graduate School, Tsinghua University, Shenzhen, China (e-mail: chen-lz25@mails.tsinghua.edu.cn).}
\IEEEcompsocthanksitem{Yujun Cai is with the University of Queensland (e-mail: vanora.caiyj@gmail.com).}
\IEEEcompsocthanksitem{Yiwei Wang and Ming-Hsuan Yang are with the University of California at Merced (e-mail: mhyang@ucmerced.edu; yiweiwang2@ucmerced.edu).}
}
}

\markboth{Journal of \LaTeX\ Class Files, January 2025}%
{Shell \MakeLowercase{\textit{et al.}}: A Sample Article Using IEEEtran.cls for IEEE Journals}


\IEEEtitleabstractindextext{

\begin{abstract}
Reliable AI systems require large language models (LLMs) to exhibit behaviors aligned with human preferences and values. However, most existing alignment approaches operate at training time and rely on additional high-quality data, incurring significant computational and annotation costs. While recent work has shown that contrastive decoding can leverage a model’s internal distributions to improve specific capabilities, its applicability remains limited to narrow behavioral scopes and scenarios. In this work, we introduce Polarity-Prompt Contrastive Decoding (PromptCD), a test-time behavior control method that generalizes contrastive decoding to broader enhancement settings. PromptCD constructs paired positive and negative guiding prompts for a target behavior and contrasts model responses—specifically token-level probability distributions in LLMs and visual attention patterns in VLMs—to reinforce desirable outcomes. This formulation extends contrastive decoding to a wide range of enhancement objectives and is applicable to both LLMs and Vision-Language Models (VLMs) without additional training. For LLMs, experiments on the “3H” alignment objectives (helpfulness, honesty, and harmlessness) demonstrate consistent and substantial improvements, indicating that post-trained models can achieve meaningful self-enhancement purely at test time. For VLMs, we further analyze contrastive effects on visual attention, showing that PromptCD significantly improves VQA performance by reinforcing behavior-consistent visual grounding. Collectively, these results highlight PromptCD as a simple, general, and cost-efficient strategy for reliable behavior control across modalities.
\end{abstract}

\begin{IEEEkeywords}
Large language models, Test-time enhancement, Contrastive decoding, Behavioral emergence, Visual attention
\end{IEEEkeywords}
}

\maketitle

\section{Introduction}
\label{sec:introduction}

\IEEEPARstart{L}{arge} language models (LLMs)~\cite{achiam2023gpt,grattafiori2024llama,yang2024qwen2} have emerged as dominant intelligent assistants, distinguished by their extensive knowledge and robust reasoning capabilities. As these systems are increasingly integrated into critical decision-making and creative workflows, ensuring their alignment with human intentions and values has become a cornerstone of trustworthy AI~\citep{bender2021dangers,gabriel2020artificial,liu2023trustworthy}. However, aligning general-purpose models is inherently complex: human preferences are diverse, context-dependent, and often conflicting across different tasks~\citep{zhang2024self,pappas2025human}. Static policies derived from standard training cannot adequately capture this variability, rendering dynamic behavioral adaptation an essential capability for next-generation systems~\citep{bhargava2023s,bi2025parameters}. Prevailing alignment efforts~\citep{yang2023alignment,tian2023fine,bi2024context} predominantly focus on the post-training stage, utilizing large curated datasets to fine-tune models via reinforcement learning from human feedback (RLHF)~\citep{ouyang2022training,bai2022training} or preference optimization~\citep{rafailov2023direct}. Despite their success, these approaches face three critical limitations: the prohibitive cost of constructing high-quality datasets; the rigidity of fixed parameters which restricts adaptability to novel contexts; and the difficulty of simultaneously optimizing conflicting objectives~\citep{guo2024controllablepreferenceoptimizationcontrollable,alami2024alignment}.
Consequently, there is a critical need for lightweight mechanisms capable of steering model behavior at inference time without the burden of retraining or altering core parameters.


Test-time steering~\cite{muennighoff2025s1,snell2024scaling} offers a complementary paradigm to address the rigidity of post-training alignment, enabling on-the-fly behavioral modulation. Existing methods primarily intervene at three stages: (1) prompt-based steering, which injects behavioral cues via natural language~\citep{sun2022recitation,zhou2023context,wei2022chain}; (2) representation-based adjustment, which modifies internal activations along learned directions~\citep{li2024inferencetimeinterventionelicitingtruthful,zhang2024truthx,kong2024aligning}; and (3) decoding-level control, which perturbs token logits during generation~\citep{li2022contrastive,chuang2023dola,shi2024trusting}. 
Despite expanding the scope of controllable alignment, these techniques exhibit notable drawbacks: prompting is often inconsistent, representation editing requires behavior-specific tuning, and decoding strategies are typically confined to predefined objectives. Developing a generalized, parameter-free test-time control mechanism that integrates multiple behavioral dimensions remains an open challenge.

In this work, we begin by investigating the efficacy of \textbf{polarity prompts}—explicit cues designed to encourage or suppress specific behaviors. 
To quantitatively ground our analysis, we take context-faithfulness as a representative case and design a \textit{Knowledge Token Capturing} algorithm to systematically examine how these cues reshape the model’s output distribution and behavioral preferences. 
Through this lens, we identify a phenomenon of \textit{latent alignment}: while polarity prompts successfully shift the probability mass toward context-faithful tokens, this internal signal frequently fails to manifest in the final generated response. The model exhibits a “willingness” to be faithful—evident in the logits captured by our algorithm—but lacks the decisiveness to override its parametric priors during standard decoding. These findings suggest that while prompting acts as a directional guide, it is insufficient for robust control , underscoring the necessity for a decoding mechanism capable of actively amplifying these latent signals.

Building on these observations, we propose \textbf{P}olarity-P\textbf{ro}mpt \textbf{C}ontrastive \textbf{D}ecoding (\textbf{PromptCD}), a unified test-time steering framework designed to enhance targeted behaviors without parameter updates. 
PromptCD employs a customizable prompting mechanism to address diverse alignment objectives, ranging from context-adherence and honesty to safety and visual grounding. For a specific target behavior, we construct paired polarity prompts that provide contrasting instructions: a positive prompt explicitly encourages the desired capability, while a negative prompt suppresses it or elicits competing tendencies, such as reliance on parametric priors or hallucinations. 
By contrasting the model responses conditioned on these cues, the framework amplifies signals specific to the target behavior while filtering out generic distribution priors. 
This approach generalizes across modalities: for general LLMs, it refines generation by contrasting token probability distributions; for Vision-Language Models (VLMs), it leverages cross-modal attention maps to improve behavior-relevant visual grounding. 
We provide an illustrative overview of this contrastive process in Figure~\ref{fig:framework_overview}, and detail the full specifications of polarity prompts for various tasks in the Appendix.

\begin{figure*}[t!]
    \centering
    \includegraphics[width=0.93\linewidth]{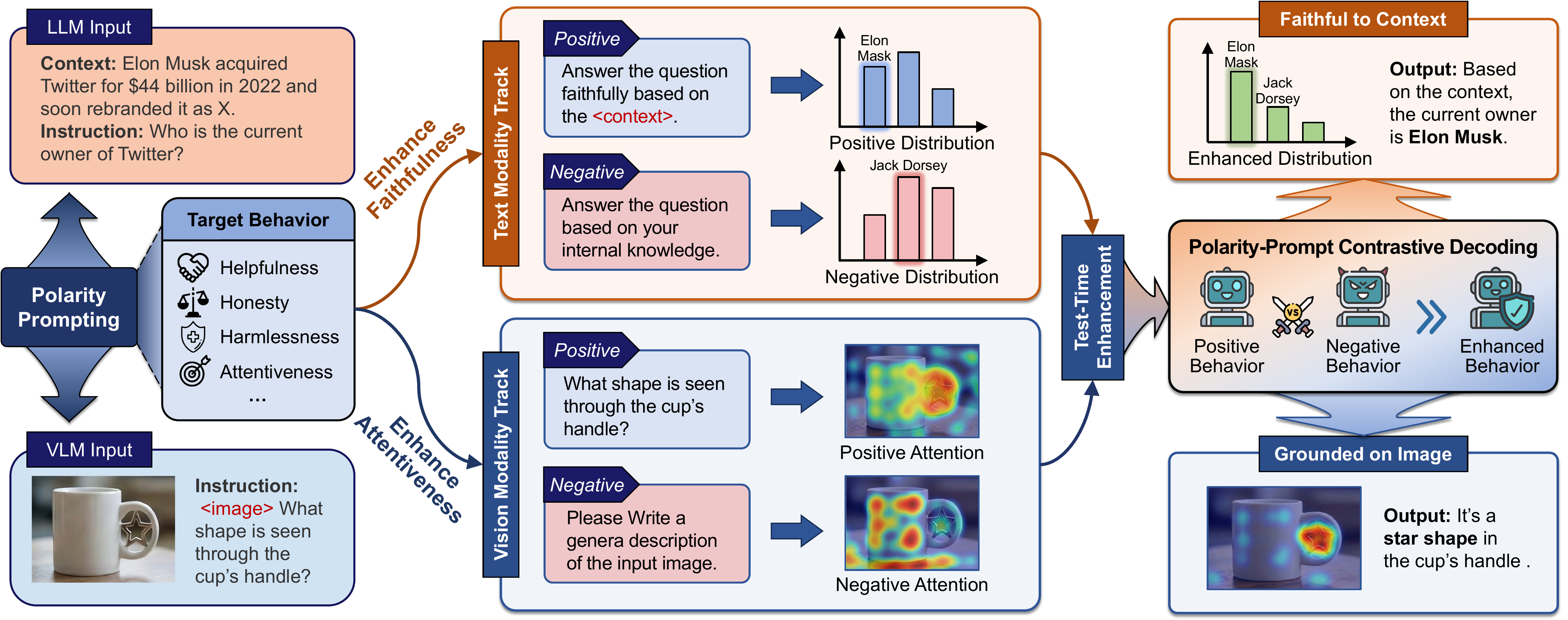}
    \caption{Overview of PromptCD. A unified framework for test-time behavior enhancement. By contrasting the logits (for text) or visual attention maps (for images) induced by positive and negative prompts, PromptCD effectively steers models toward desired goals without retraining. We demonstrate this capability through faithfulness enhancement in LLMs and visual attentiveness improvement in VLMs.}
    \label{fig:framework_overview}
\end{figure*}

To evaluate the effectiveness and generality of our approach, we conduct comprehensive cross-modal experiments encompassing both Large Language Models (LLMs) and Vision-Language Models (VLMs). For LLMs, we assess behavioral alignment across the “3H” dimensions: helpfulness, honesty, and harmlessness~\cite{bai2022constitutionalaiharmlessnessai}. Specifically, helpfulness is evaluated via context-faithfulness tasks on benchmarks such as NQ~\citep{kwiatkowski2019natural}, ConFiQA~\citep{bi2024context}, and CoConflictQA~\citep{huang2025pip}; honesty is measured using TruthfulQA~\citep{lin2021truthfulqa} and FactScore~\citep{min2023factscore}; and harmlessness is tested on SafeEdit~\citep{wang2024detoxifying}. For VLMs, we extend PromptCD to Visual Question Answering (VQA). Here, we analyze not only token probability shifts but also the modulation of visual attention, demonstrating that polarity prompts sharpen the model's focus on semantically relevant regions to improve answer grounding. Complementing these quantitative metrics, we provide granular token-level and attention-based interpretability analyses, elucidating how PromptCD dynamically reshapes generation trajectories during decoding. Empirical results confirm that PromptCD consistently enhances alignment across modalities, establishing it as a lightweight, parameter-free mechanism for flexible test-time behavior control.

\section{Preliminaries}
\label{sec:preliminaries}

In this section, we formalize the generation processes of Large Language Models (LLMs) and Vision-Language Models (VLMs). This formulation establishes the theoretical basis for our proposed method, which intervenes at the prediction level for LLMs and the attention level for VLMs.

\subsection{Autoregressive Generation in LLMs}
\label{sec:prelim_llm}

Large language models function as probabilistic estimators over sequences of tokens. 
Given a sequence of previous tokens $\mathcal{X}_{<t} = \{x_1, x_2, \ldots, x_{t-1}\}$, the model aims to predict the next token $x_t$ by estimating the conditional probability distribution. 
Formally, an LLM parameterized by $\theta$ computes this probability via a softmax output layer:

\begin{equation}
    \text{I\kern-0.15em P}_\theta(x_t|\mathcal{X}_{<t}) = 
    \mathrm{softmax}\!\left(\phi(\mathbf{h}_t)\right),
\label{eq:lm_prob}
\end{equation}

where $\mathbf{h}_t$ denotes the contextualized hidden state at step $t$, and $\phi(\cdot)$ is a linear transformation projecting $\mathbf{h}_t$ into the vocabulary space to produce logits.
The hidden state $\mathbf{h}_t$ is derived through a stack of Transformer layers, processing the embedded sequence $\mathbf{X}_{<t}$ via self-attention and feed-forward operations:

\begin{equation}
    \mathbf{h}_t = \mathrm{FFN}\!\left(\mathrm{Attn}(\mathbf{X}_{<t})\right).
\label{eq:lm_hidden}
\end{equation}

The attention module $\mathrm{Attn}(\cdot)$ captures contextual dependencies across tokens, and the feed-forward network $\mathrm{FFN}(\cdot)$ applies nonlinear transformations to produce the contextualized representation.
During decoding, the model autoregressively samples the next token $x_t$ according to the predicted probability distribution $\text{I\kern-0.15em P}_\theta(x_t|\mathcal{X}_{<t})$.
While decoding strategies (e.g., greedy search, nucleus sampling) determine the specific selection rule, the generation trajectory is fundamentally governed by this probability distribution.
Consequently, our proposed PromptCD method for LLMs operates directly on Equation~\ref{eq:lm_prob}, modulating the output logits to amplify favorable behavioral signals before sampling.

\subsection{Visual-Textual Grounding in VLMs}
\label{sec:prelim_vlm}

VLMs extend the autoregressive paradigm by incorporating an input image $\mathcal{I}$ into the generation process.
The computation of hidden states relies on a multimodal attention mechanism that jointly attends to textual history and encoded visual features $\mathbf{V}$:

\begin{equation}
    \mathbf{h}_t = \mathrm{FFN}\!\left(\mathrm{Attn}_{\text{text}+\text{vision}}(\mathbf{X}_{<t}, \mathbf{V})\right),
\label{eq:vlm_hidden}
\end{equation}

where $\mathrm{Attn}_{\text{text}+\text{vision}}(\cdot)$ computes attention weights that quantify the cross-modal relevance of specific visual regions in $\mathbf{V}$.
The final probability is similarly computed as:

\begin{equation}
    \text{I\kern-0.15em P}_\theta(x_t|\mathcal{X}_{<t}, \mathcal{I}) = 
    \mathrm{softmax}\!\left(\phi(\mathbf{h}_t)\right).
\label{eq:vlm_prob}
\end{equation}

While the probabilistic output (Equation~\ref{eq:vlm_prob}) remains the final decision layer, the inclusion of cross-modal attention in Equation~\ref{eq:vlm_hidden} introduces a new dimension of controllability.

\begin{figure*}[t!]
    \centering
    \includegraphics[width=0.95\linewidth]{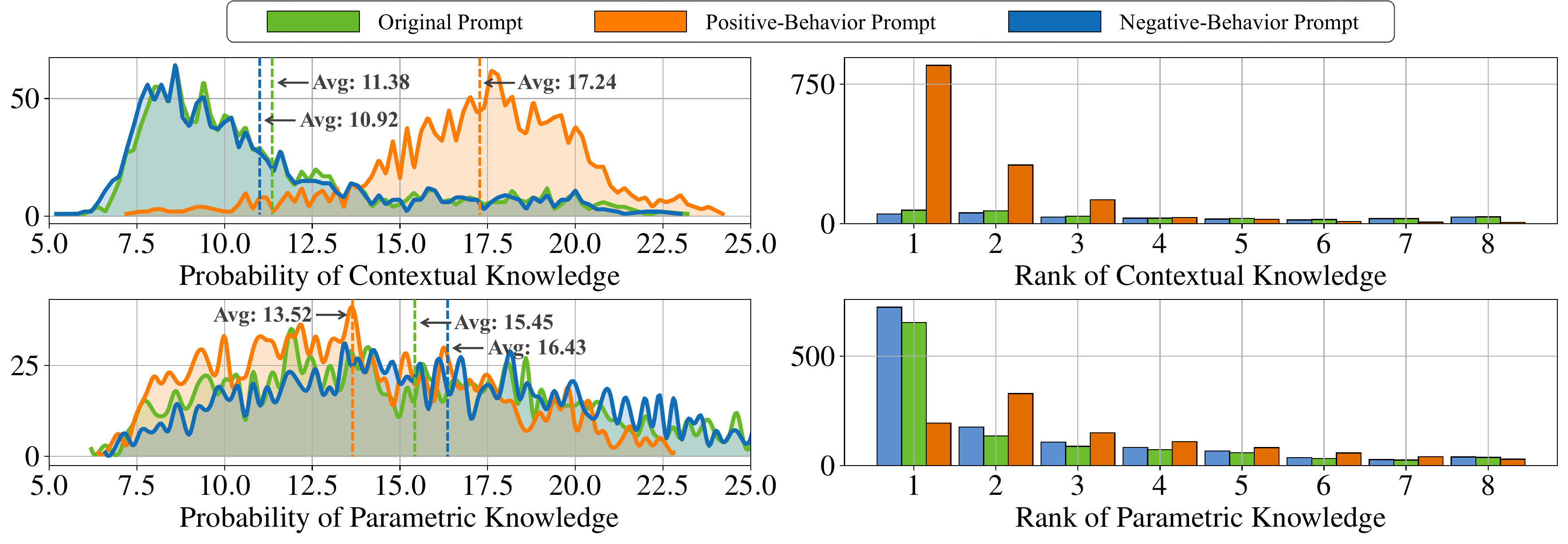}
    \vspace{-4mm}
    \caption{Comparison of contextual and parametric knowledge shifts under positive and negative prompts.  The first discriminative tokens are identified to represent the corresponding knowledge, and their logit and rank distributions are tracked to illustrate the underlying distributional dynamics.}
    \label{fig:pre_stat}
\end{figure*}

Just as LLMs conditioned on positive ($P$) and negative ($N$) prompts yield distinct probability distributions $\mathbb{P}^{(P)}$ and $\mathbb{P}^{(N)}$, VLMs conditioned on these prompts exhibit distinct visual attention patterns.
We define $A_{l}^{(P)}(\mathcal{I})$ as the attention distribution at layer $l$ under the positive prompt, which captures behavior-specific positive signals (e.g., semantic relevance to the question).
Conversely, we define $A_{l}^{(N)}(\mathcal{I})$ as the attention under the negative prompt, which primarily reflects generic negative patterns (e.g., visual saliency or noise) shared across contexts.
PromptCD leverages this analogy by contrasting these attention maps—$A_{l}^{(P)}(\mathcal{I})$ versus $A_{l}^{(N)}(\mathcal{I})$—to isolate and amplify the task-relevant visual grounding signals, similar to how it contrasts token probabilities in LLMs.

\section{Empirical Analysis: How Polarity Prompts Reshape Output Distributions}
\label{sec:explore}


LLMs exhibit significant behavioral variability conditioned on their input prompts. 
To better understand how prompts explicitly steering toward or away from a target behavior influence model generation, we perform an exploratory study on \textit{polarity prompts}. 
This section formalizes the definition of polarity prompts, investigates their effects through the lens of \textit{context-faithfulness}, introduces a token-level analysis algorithm, and highlights a key challenge in behavioral emergence.

\subsection{Definition of Polarity Prompts}
\label{sec:polarity_def}

We conceptualize behavioral steering as a modulation process driven by paired opposing cues.
Rather than treating prompts merely as natural language instructions, we define Polarity Prompts as paired constraints designed to induce distinct behavioral distributions.
Given a target behavior $B$, we construct a pair of prompts $(P, N)$:

\begin{itemize}
    \item Positive Prompt ($P$): An instruction designed to explicitly encourage the expression of the target behavior $B$ (e.g., faithful reasoning) within the generation context.
    \item Negative Prompt ($N$): An instruction designed to suppress $B$ or explicitly elicit a competing behavioral tendency (e.g., reliance on parametric priors).
\end{itemize}

For the specific case of \textit{context-faithfulness} in LLMs, the goal is to guide the model to reason based on provided contextual information rather than its pre-trained memory.
Thus, a positive prompt might be: \textit{“Please answer the question faithfully based on the provided context.”}
Conversely, the negative prompt actively encourages the competing behavior: \textit{“Please answer the question based on your own parametric knowledge.”}

Formally, given an input instruction or question $X_q$, we concatenate it with either a positive prompt $X_p$ or a negative prompt $X_n$. 
The conditional probability of generating the next token under a positive prompt is formulated as:

\begin{equation}
    \text{I\kern-0.15em P}(x^{(P)}_m|x^{(P)}_{<m}) = 
    \mathrm{softmax}(\phi(\mathbf{h}^{(P)}_m)),
\label{eq:pe}
\end{equation}

where $\mathbf{h}^{(P)}_m$ is the hidden representation of the sequence $X_q + X_p$.
A corresponding distribution $\text{I\kern-0.15em P}(x^{(N)}_n|x^{(N)}_{<n})$ is similarly derived from the sequence $X_q + X_n$.
By contrasting these two distributions, we can rigorously quantify how polarity prompts modulate the model’s internal generation dynamics and shift token preferences.

\subsection{Quantifying Behavior Shifts via Token-Level Trace}
\label{sec:token_analysis}

To empirically verify whether polarity prompts effectively steer the model, we adopt \textit{context-faithfulness} as a representative testbed.
We utilize the \textsc{MQuAKE} benchmark~\citep{zhong2023mquake}, which consists of questions where the ground truth (Parametric Knowledge) conflicts with a provided counterfactual context (Contextual Knowledge).
For instance, given the context \textit{"Paris is the capital of the United States"}, a faithful model should answer \textit{"United States"}, while a model relying on priors will answer \textit{"France"}.

Standard evaluation metrics (e.g., accuracy) only reveal the final discrete output, masking the internal dynamics.
To bridge this gap, we propose a fine-grained Knowledge Token Capturing Algorithm (Algorithm~\ref{alg:alg}).
This algorithm acts as a probe to trace the competition between the two knowledge sources at the token level.
Specifically, it scans the generated logits to identify discriminative tokens—the first token position where the contextual answer string diverges from the parametric answer string (e.g., the token "States" vs. "France").
By tracking the probability mass assigned to these specific tokens under different prompting conditions, we can rigorously quantify the model's internal preference shift, even if the final output remains unchanged.

\begin{algorithm}
\caption{Knowledge Token Capturing}
\label{alg:alg}
\begin{algorithmic}[1]
\Require The LLM generates a token sequence of length $n$, 
$\mathcal{V}$: vocabulary of the LLM, 
$\mathcal{P}_i \in (\mathcal{P}_1, \mathcal{P}_2, \ldots, \mathcal{P}_n)$: the logits distribution for each token, 
$S_{\mathrm{cont}}$: string of the contextual answer (from the counterfactual context), 
$S_{\mathrm{para}}$: string of the parametric answer (ground truth).
\Ensure Captured contextual knowledge logits $P_{\mathrm{cont}}$ and parametric knowledge logits $P_{\mathrm{para}}$.

\State Initialize $P_{\mathrm{cont}} \gets \textit{None}$, $P_{\mathrm{para}} \gets \textit{None}$
\State $S_{\text{com}} \gets \text{COM}(S_{\mathrm{cont}}, S_{\mathrm{para}})$ 
// common substrings
\For{$\mathcal{P}_i \in (\mathcal{P}_1, \mathcal{P}_2, \ldots, \mathcal{P}_n)$}
    \State Let $x_i \gets \arg\max \mathcal{P}_i$ and $x_i' \gets \text{Decode}(x_i)$.
    \If{$x_i' \notin S_{\mathrm{cont}}$ \textbf{and} $x_i' \notin S_{\mathrm{para}}$}
        \State \textbf{continue} 
    \EndIf
    \For{each token $x_j \in \mathcal{V}$ (descending order by $P_{i,j}$)}
        \State Decode $x_j$ into string $x'_j$.
        \State \textbf{if} {$x'_j \in S_{\text{com}}$ \textbf{and} $P_{\mathrm{cont}} = P_{\mathrm{para}} = \textit{None}$: }
        \textbf{break} 
        \State \textbf{if} {$x'_j \in S_{\mathrm{cont}}$ \textbf{and} $P_{\mathrm{cont}} = \textit{None}$: }
             $P_{\mathrm{cont}} \gets P_{i,j}$
        \State \textbf{if} {$x'_j \in S_{\mathrm{para}}$ \textbf{and} $P_{\mathrm{para}} = \textit{None}$: }
            $P_{\mathrm{para}} \gets P_{i,j}$
    \EndFor
\EndFor\\
\Return $P_{\mathrm{cont}}$, $P_{\mathrm{para}}$
\end{algorithmic}
\end{algorithm}

As visualized in Figure~\ref{fig:pre_stat}, our analysis yields striking observations regarding the shifts in probability distributions and token rankings. 
With positive prompts, the probability distribution of contextual knowledge shifts significantly to the right, whereas the probability of parametric knowledge remains largely stable or exhibits a minor decline. 
Conversely, negative prompts tend to increase the probability of parametric knowledge but exert negligible influence on contextual knowledge, effectively rendering the context invisible. 
Critically, this distributional shift directly impacts behavioral realization: under positive prompts, there is a substantial surge in the frequency of contextual knowledge achieving Rank 1, thereby enabling the emergence of context-faithful behavior. 
In contrast, negative prompts reinforce the dominance of parametric outputs, solidifying the model's reliance on internal priors.

\begin{figure*}[t!]
    \centering
    \includegraphics[width=0.98\linewidth]{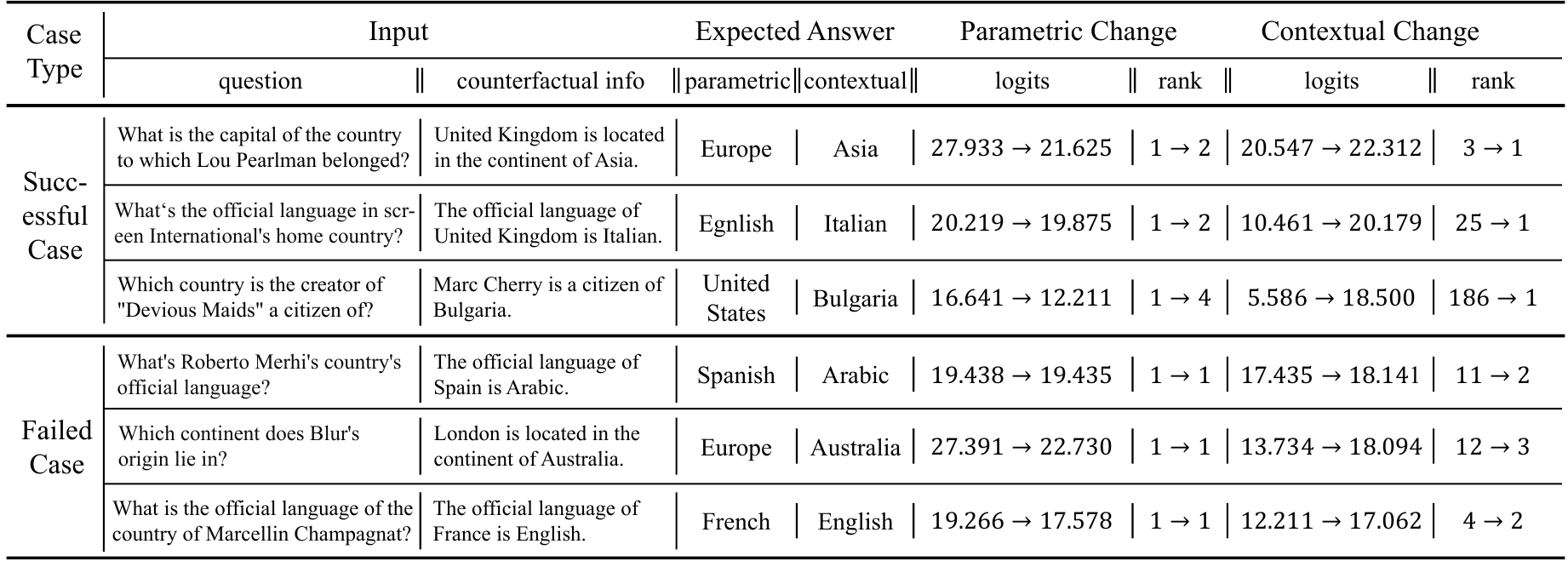}
    \caption{Selected cases showing changes in the first token for both parametric and contextual knowledge. The cases are obtained by introducing positive prompts carrying counterfactual information into the \llamaa model. '$\rightarrow$' denotes the knowledge shift after incorporating positive prompts. 'Logits' and `rank' refer to the first discriminative token in the knowledge answer, indicating the model's knowledge confidence.}
    \label{fig:pre_case}
\end{figure*}

\subsection{Challenge of Behavioral Emergence}

Despite the clear distributional shift observed above, we identify a persistent phenomenon that hinders practical alignment: \textbf{Latent Alignment} does not guarantee Behavioral Emergence.
While positive prompts significantly boost the confidence of LLMs regarding new knowledge derived from context, we find that there are still frequent instances where this context knowledge ranks prominently but fails to secure the Top-1 position, as illustrated in Figure~\ref{fig:pre_stat}.

We term this phenomenon \textbf{"stubborn knowledge"}, referring to scenarios where faithful behavior fails due to either excessive confidence in existing parametric priors or insufficient confidence in the newly introduced knowledge.
The cases in Figure~\ref{fig:pre_case} deeply reveal the failure pattern of faithful behavioral emergence under positive prompting in addressing stubborn knowledge.
Specifically, failure often occurs when an extremely small gap remains between the target and the parametric knowledge, despite the significant increase in new knowledge logits induced by the positive prompt.
Taking the last case in Figure~\ref{fig:pre_case} as an example, although the logit for the faithful answer ("English") rises sharply, it still lags behind the entrenched parametric prior ("French") by a distinct margin of 0.516 in terms of logit distribution.

Under standard decoding strategies such as greedy search, the model fails to capitalize on this increased probability and instead selects the parametric token, resulting in an unfaithful output.
This finding highlights a critical limitation of standard prompting: it functions as a probabilistic bias rather than a deterministic control mechanism.
Although the model demonstrates a distinct latent tendency towards faithfulness (evidenced by elevated logits), it often lacks sufficient probability mass to surpass the decision boundary imposed by robust parametric priors.
This misalignment between latent representation and final behavior serves as the primary motivation for our work.
It underscores the necessity for a decoding-time intervention capable of actively amplifying these latent probability shifts, thereby transforming a tentative internal tendency into a stable, manifest behavioral realization.

\section{Polarity-Prompt Contrastive Decoding}

Motivated by the findings in Section \ref{sec:explore}, we propose Polarity-Prompt Contrastive Decoding (PromptCD), a unified test-time steering framework designed to flexibly enhance targeted behaviors in LLMs. The core principle of PromptCD is to leverage polarity prompts—paired instructions that explicitly encourage (positive prompt) or discourage (negative prompt) a desired behavior—to modulate the model's inference dynamics.
Specifically, PromptCD isolates and amplifies the signals aligned with the target behavior by contrasting the model's outputs conditioned on these opposing cues. This mechanism is versatile across modalities: for LLMs, it operates by contrasting token probability distributions to refine generation logits; for VLLMs, it extends to contrasting cross-modal attention maps to highlight behavior-relevant visual regions. We detail these two distinct implementations in the following subsections.

\begin{figure}[t]
    \centering
    \includegraphics[width=\linewidth]{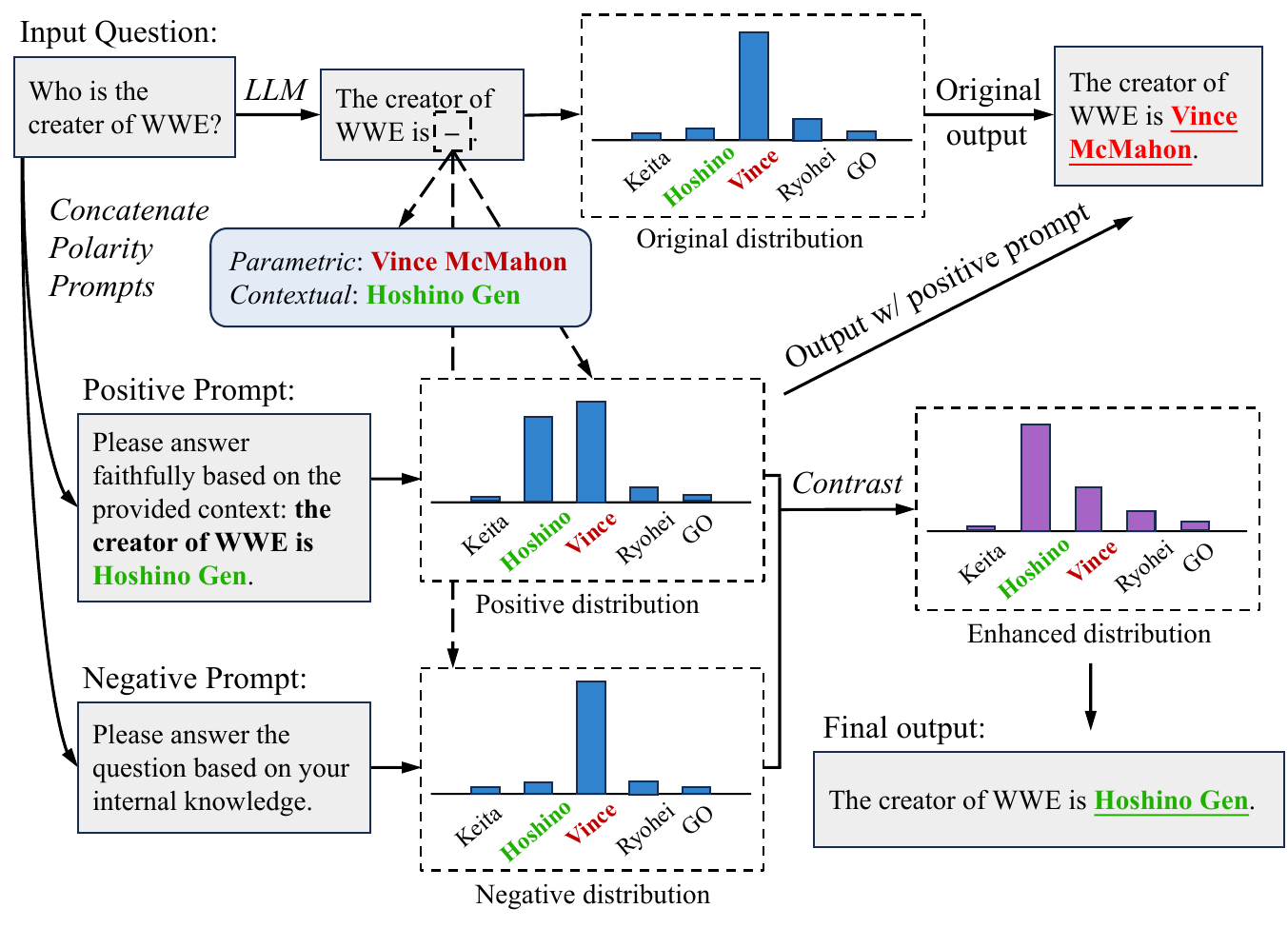}
    \vspace{-6mm}
    \caption{Illustration of PromptCD enhancing context-faithfulness.  By contrasting the probability distributions induced by positive and negative prompts, PromptCD amplifies contextual signals, guiding the LLM to generate faithful responses to the input question.}
    \vspace{-4mm}
    \label{fig:framework_llm}
\end{figure}

\subsection{Contrastive Decoding via Token Logits}
\label{sec:logits}

Figure~\ref{fig:framework_llm} conceptually illustrates the PromptCD workflow applied to LLMs. 
The fundamental intuition behind our approach is that the probability distribution generated by a large language model is a composition of task-specific reasoning and generic parametric priors. 
When conditioned on a \textit{positive prompt} (e.g., specific instructions to be faithful), the output distribution $\mathbb{P}^{(P)}$ contains both the desired behavioral signals and the model's inherent biases. 
Conversely, when conditioned on a \textit{negative prompt} (e.g., instructions to rely on internal knowledge or generic chitchat), the output distribution $\mathbb{P}^{(N)}$ predominantly captures these inherent biases, hallucinations, or context-ignoring tendencies.
PromptCD aims to distill the pure behavioral signal by contrastively subtracting the latter from the former.

Formally, let $\mathbb{P}(x_t | x_{<t})$ denote the model's predicted probability for the next token $x_t$ given the preceding context $x_{<t}$.
At each decoding step $t$, we instantiate two parallel inference contexts: the sequence augmented with the positive prompt, denoted as $X^{(P)}$, and the sequence with the negative prompt, $X^{(N)}$.
The model independently computes the conditional probability distributions for the next token under both settings: 
$\mathbb{P}^{(P)}(x_t) = \mathbb{P}(x_t | X^{(P)}_{<t})$ and $\mathbb{P}^{(N)}(x_t) = \mathbb{P}(x_t | X^{(N)}_{<t})$.

\subsubsection{Contrastive Scoring Function.}
To effectively steer the generation toward the target behavior, we employ a contrastive scoring objective that amplifies tokens unique to the positive context while penalizing those that are highly probable in the negative context.
The adjusted probability distribution $\hat{\mathbb{P}}(x_t)$ is derived by normalizing the contrastive scores via a softmax operation:

\begin{equation}
    \hat{\mathbb{P}}(x_t) = \mathrm{softmax}\left( \mathcal{F}\left( \mathbb{P}^{(P)}(x_t), \mathbb{P}^{(N)}(x_t) \right) \right),
    \label{eq:promptcd}
\end{equation}

where the operator $\mathcal{F}(\cdot, \cdot)$ computes the token-level logits modification.
Drawing inspiration from Contrastive Decoding~\citep{li2022contrastive}, we formulate this operator as a weighted subtraction in the log-probability space. 
This operation is mathematically equivalent to maximizing the mutual information between the token and the positive control signal, while minimizing the influence of the negative prior:

\begin{equation}
    \mathcal{F}(\cdot) = 
    \begin{cases} 
    \log \mathbb{P}^{(P)}(x_t) - \gamma \cdot \log \mathbb{P}^{(N)}(x_t), & \text{if } x_t \in \mathcal{V}_{\text{head}}, \\
    -\infty, & \text{otherwise},
    \end{cases}
    \label{eq:scoring}
\end{equation}

where $\gamma \geq 0$ is a critical hyperparameter serving as the \textit{contrastive coefficient}. 
This coefficient controls the strength of the penalty applied to the negative distribution. 
Intuitively, for high-frequency stop words or generic tokens where $\mathbb{P}^{(P)} \approx \mathbb{P}^{(N)}$, the subtraction significantly dampens their scores. 
In contrast, for discriminative tokens that align strictly with the target behavior (high $\mathbb{P}^{(P)}$ and low $\mathbb{P}^{(N)}$), the score is preserved or boosted.

\subsubsection{Adaptive Plausibility Constraint (APC).}
While contrastive scoring effectively highlights discriminative features, it introduces a risk of the ``amateur correction'' pathology. 
Specifically, a token that is extremely unlikely under the negative prompt (very large negative $\log \mathbb{P}^{(N)}$) might receive an artificially inflated score after subtraction, even if it is nonsensical in the positive context.
To mitigate this and ensure the semantic fluency of the generated text, we adopt the Adaptive Plausibility Constraint (APC)~\citep{li2022contrastive}. 
This constraint imposes a dynamic validity mask, restricting the candidate pool $\mathcal{V}_{\text{head}}$ to only those tokens that are already plausible under the positive prompt:

\begin{equation}
    \mathcal{V}_{\text{head}} = \left\{ x_t \in \mathcal{V} : \mathbb{P}^{(P)}(x_t) \geq \lambda \max_{w \in \mathcal{V}} \mathbb{P}^{(P)}(w) \right\},
    \label{eq:apc}
\end{equation}

where $\lambda \in (0, 1]$ is a filtering threshold. 
By setting the scores of tokens outside this set to $-\infty$, we ensure that the optimization landscape remains grounded within the valid semantic manifold of the task-compliant model, preventing the generation of gibberish.

\subsubsection{Synchronized Dual-Sequence Decoding.}
A distinct structural feature of PromptCD, which differentiates it from standard re-ranking or layer-contrastive methods~\citep{chuang2023dola}, is the simultaneous maintenance of two synchronized generation trajectories. 
Although the generation is guided by the contrastive distribution, the resulting token must maintain coherence in both prompt contexts to allow for iterative contrast.
Let $x^*_t$ be the token sampled from the adjusted distribution $\hat{\mathbb{P}}(x_t)$ (Eq.~\ref{eq:promptcd}). 
Crucially, this single selected token is appended to \textit{both} the positive and negative context buffers before proceeding to step $t+1$:

\begin{equation}
    X^{(P)}_{<t+1} \leftarrow X^{(P)}_{<t} \oplus x^*_t, \quad 
    X^{(N)}_{<t+1} \leftarrow X^{(N)}_{<t} \oplus x^*_t.
\end{equation}

This strict synchronization ensures that $\mathbb{P}^{(P)}$ and $\mathbb{P}^{(N)}$ are always conditioned on identical generated histories. 
Without this synchronization, the two streams would rapidly diverge into semantically disjoint trajectories, rendering the contrastive subtraction meaningless. 
This mechanism allows PromptCD to correct the model's behavior token-by-token in real-time, effectively steering the entire generation path.

\subsection{Contrastive Decoding on Cross-Modal Attention}

Similar to LLMs, VLMs also face the challenge of behavioral alignment, particularly in how they attend to visual information.
Visual noise often distracts the model from task-relevant regions, leading to hallucinations or unfaithful responses.
While token-level contrastive decoding effectively enhances textual generation, VLMs introduce an additional modality where PromptCD can be naturally extended: visual attention patterns.
In visual tasks, we define the target \textit{behavior} as the VLM's capability to precisely attend to task-relevant image regions.
To steer this behavior, we introduce paired polarity prompts: a \textbf{positive prompt}, corresponding to a task-specific question, encourages the model to focus on specific regions required for reasoning; conversely, a \textbf{negative prompt}, corresponding to a general instruction, elicits a broad, behavior-agnostic attention pattern over the entire image.
As established in Equation~\ref{eq:vlm_hidden}, VLMs jointly attend to textual tokens and visual features through multimodal attention $\mathrm{Attn}_{\text{text}+\text{vision}}(\mathbf{X}_{<t}, \mathbf{V})$.
We observe that polarity prompts not only modulate token probability distributions but also reshape the spatial allocation of visual attention—a phenomenon we leverage to enhance visual grounding in vision-language tasks.

\begin{figure}[t]
    \centering
    \includegraphics[width=\linewidth]{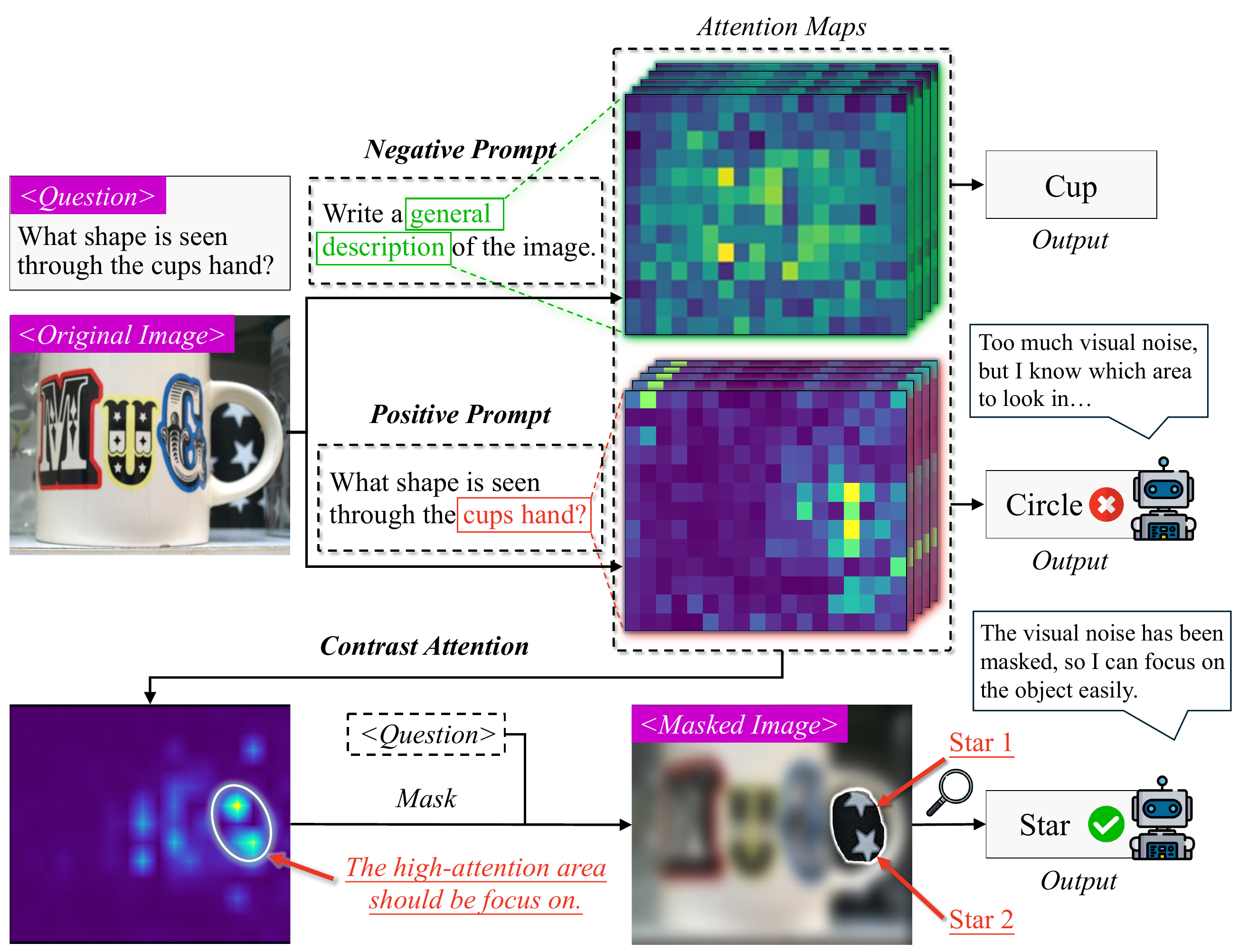}
    \vspace{-6mm}
    \caption{The PromptCD framework for cross-modal attention refinement.  By contrasting the task-specific attention distribution $A^{(P)}$ (derived from positive prompts) against the general attention pattern $A^{(N)}$ (derived from negative prompts), the method computes a refined attention map $\hat{A}$.  This contrastive mechanism effectively isolates and highlights request-relevant visual regions, thereby enhancing visual grounding.}
    \vspace{-4mm}
    \label{fig:vlm_framework}
\end{figure}

\subsubsection{Visual Attention Decomposition.}
As Equation~\ref{eq:scoring} contrasts token logits, we contrast attention distributions over visual regions to decompose cross-modal attention into behavior-specific positive signals and generic negative patterns.
Recall from Section~\ref{sec:prelim_vlm} that $A_{l}^{(P)}(\mathcal{I})$ and $A_{l}^{(N)}(\mathcal{I})$ represent attention distributions at layer $l$ under positive and negative prompts, respectively.
We model the positive attention as a compositional interaction between the behavior-specific signal $\mathcal{F}^{(+)}(P, \mathcal{I})$ induced by the positive prompt and the generic visual pattern $\mathcal{F}^{(-)}(\mathcal{I})$ shared across prompts:
\begin{equation}
A_{l}^{(P)}(\mathcal{I}) = \mathcal{F}^{(-)}(\mathcal{I}) \otimes \mathcal{F}^{(+)}(P, \mathcal{I}),
\label{eq:vlm_attn_decomp}
\end{equation}
where $\otimes$ denotes the Hadamard product.
Under the negative prompt, which provides only generic cues (e.g., \textit{"Write a general description of the image"}), the behavior-specific signal becomes approximately uniform, causing attention to primarily reflect the generic pattern: $A_{l}^{(N)}(\mathcal{I}) \approx \mathcal{F}^{(-)}(\mathcal{I})$.

\subsubsection{Contrastive Visual Attention}
To isolate the behavior-specific positive signal $\mathcal{F}^{(+)}(P, \mathcal{I})$, we compute a contrastive attention map $\hat{A}_{l}$ by normalizing the positive attention with respect to the negative attention:

\begin{equation}
\hat{A}_{l,i} = \frac{A_{l,i}^{(P)}}{A_{l,i}^{(N)} + \lambda},
\label{eq:vlm_contrast_attn}
\end{equation}

where $A_{l,i}^{(P)}$ and $A_{l,i}^{(N)}$ denote the attention weights at the $i$-th visual token in layer $l$ under positive and negative prompts respectively, and $\lambda > 0$ is a regularization parameter.
Substituting the decomposition from Equation~\ref{eq:vlm_attn_decomp}, we obtain:

\begin{equation}
\hat{A}_{l,i} \approx \frac{\mathcal{F}^{(-)}_{i}(\mathcal{I}) \cdot \mathcal{F}^{(+)}_{i}(P, \mathcal{I})}{\mathcal{F}^{(-)}_{i}(\mathcal{I}) + \lambda} \approx \mathcal{F}^{(+)}_{i}(P, \mathcal{I}),
\label{eq:vlm_contrast_approx}
\end{equation}

where the approximation holds when $\mathcal{F}^{(-)}_{i}(\mathcal{I}) \gg \lambda$.
Thus, the contrastive operation cancels out the generic negative pattern $\mathcal{F}^{(-)}(\mathcal{I})$, retaining only the behavior-specific positive signal $\mathcal{F}^{(+)}(P, \mathcal{I})$—analogous to how Equation~\ref{eq:scoring} isolates behavior-favorable tokens in LLMs.

\subsubsection{Application to Visual Question Answering.}
Having obtained the semantically refined contrastive attention maps $\{\hat{A}_{l}\}$ as illustrated in Figure~\ref{fig:vlm_framework}, we now apply PromptCD to enhance visual grounding in VQA tasks.
Positive prompts encourage focused visual reasoning over specific image regions, whereas negative prompts elicit generic scene descriptions without semantic focus.
Since different layers capture complementary visual information, we fuse contrastive attention maps across the layer range $\mathcal{L}$ through weighted aggregation, where deeper layers receive higher fusion weights as they encode more refined semantic information.
Request-relevant visual regions are then identified by applying a top-$p$ percentile threshold $\tau = \mathcal{Q}_p(S)$ to the fused attention map $S$, which retains the top $p \in (0,1]$ proportion of pixels.
Connected component analysis extracts coherent regions from the thresholded map, and we select the top-$K$ regions ranked by cumulative attention scores to generate the enhanced image through $\mathcal{I}_{\text{refined}} = \Phi(\mathcal{I}, M^*)$, where $\Phi$ applies masking, cropping, and resizing operations, and $K$ controls the maximum number of regions to preserve.

\section{Experimental Methodology}

\subsection{Setup for Text-Only Models}

We evaluate our PromptCD in terms of the “3H” dimensions for text-only LLMs to assess its effectiveness in test-time behavior enhancement.
Our evaluation is conducted on four popular open-source LLMs: \llamaa~\citep{touvron2023llama}, \llamac~\citep{grattafiori2024llama}, \Mistral~\citep{jiang2023mistral7b}, and \textsc{Qwen2.5-7B-instruct}~\citep{yang2024qwen2}.

\subsubsection{Helpfulness} 

Helpfulness of LLMs requires the model to faithfully follow human-provided contextual instructions. 
To this end, we conduct experiments using the Natural Questions (NQ)~\citep{kwiatkowski2019natural}, ConFiQA~\citep{bi2024context}, and CoConflictQA~\citep{huang2025pip} datasets.
In NQ, we further utilize the counterfactual information provided by the dataset. All datasets simulate knowledge conflicts through counterfactual contexts, aiming to examine whether the model can still generate helpful responses according to the given context when external instructions contradict its internal parametric knowledge.
We use ConR (recall of context) and ParR (recall of parameters) as evaluation metrics. ConR measures whether the generated responses align with the provided context, while ParR evaluates their alignment with the model’s internal knowledge. Specifically, we also adopt the memorization ratio $\text{MR} = \frac{\text{ParR}}{\text{ParR} + \text{ConR}}$, which captures the tendency to favor parameters over context.
We use two prompt-based baselines: the Attributed (Attr) and the Opinion-based (Opin) prompt~\citep{zhou2023context}.

\begin{table*}[t!]
    \centering
    \caption{Performance (\%) of PromptCD on LLM helpfulness evaluation.}
    \vspace{-2mm}
    \renewcommand{\arraystretch}{1.2}
    \renewcommand{\arraystretch}{1}
    \resizebox{\textwidth}{!}{
    \begin{tabular}{l|l|ccc|ccc|ccc}
        \toprule
        \multirow{2}{*}{\textbf{Model}} & \multirow{2}{*}{\textbf{Method}} & \multicolumn{3}{c|}{\textbf{NQ}} & \multicolumn{3}{c|}{\textbf{ConFiQA}} & \multicolumn{3}{c}{\textbf{CoConflictQA}} \\
        \cmidrule(lr){3-5} \cmidrule(lr){6-8} \cmidrule(lr){9-11}
        & & \text{ConR}$(\uparrow)$ & \text{ParR}$(\downarrow)$ & \text{MR}$(\downarrow)$ & \text{ConR}$(\uparrow)$ & \text{ParR}$(\downarrow)$ & \text{MR} $(\downarrow)$ & \text{ConR}$(\uparrow)$ & \text{ParR}$(\downarrow)$ & \text{MR}$(\downarrow)$ \\
        \midrule 
         \multirow{4}{*}{\textsc{LLaMA2-7B-chat}} & \cellcolor{gray!20}Vanilla & \cellcolor{gray!20}43.35 & \cellcolor{gray!20}43.75 & \cellcolor{gray!20}50.24   & \cellcolor{gray!20}69.66 & \cellcolor{gray!20}28.14 & \cellcolor{gray!20}28.78   & \cellcolor{gray!20}62.07 & \cellcolor{gray!20}18.76 & \cellcolor{gray!20}23.20   \\
         & Attr & 64.39 & 21.99 & 25.46 & 82.03 & 18.76 & 18.61 & 65.46 & 13.97 & 17.58 \\
        & Opin & 74.59 & 15.39 & 17.11 & 88.62 & 15.38 & 14.79 & 67.26 & 13.57 & 16.79 \\
        & PromptCD & \textbf{75.99} & \textbf{6.39} & \textbf{7.76} & \textbf{90.42} & \textbf{13.76} & \textbf{13.21} & \textbf{70.25} & \textbf{11.16} & \textbf{13.71} \\
        \midrule
        \multirow{4}{*}{\textsc{LLaMA3-8B-instruct}} & \cellcolor{gray!20}Vanilla & \cellcolor{gray!20}43.92 & \cellcolor{gray!20}34.05 & \cellcolor{gray!20}43.47   & \cellcolor{gray!20}54.23 & \cellcolor{gray!20}22.39 & \cellcolor{gray!20}29.22   & \cellcolor{gray!20}68.46 & \cellcolor{gray!20}29.74 & \cellcolor{gray!20}30.29   \\
        & Attr & \textbf{74.39} & 41.39 & 35.75 & 87.82 & 37.52 & 29.93 & 76.24 & 22.35 & 22.67 \\
        & Opin & 69.99 & 20.99 & 23.07 & \textbf{91.21} & 22.95 & 20.10 & 73.85 & 18.56 & 20.08 \\
        & PromptCD & 72.69 & \textbf{6.59} & \textbf{8.33} & 90.92 & \textbf{19.56} & \textbf{17.70} & \textbf{82.43} & \textbf{15.76} & \textbf{16.05} \\
        \midrule
        \multirow{4}{*}{\textsc{Mistralv0.3-7B-instruct}} & \cellcolor{gray!20}Vanilla & \cellcolor{gray!20}46.19 & \cellcolor{gray!20}58.59 & \cellcolor{gray!20}55.92   & \cellcolor{gray!20}64.68 & \cellcolor{gray!20}25.87 & \cellcolor{gray!20}28.57   & \cellcolor{gray!20}67.06 & \cellcolor{gray!20}19.96 & \cellcolor{gray!20}22.94   \\
        & Attr & 71.59 & 25.59 & 26.34 & 79.64 & 18.56 & 18.90 & 67.07 & 16.37 & 19.62 \\
        & Opin & 73.39 & 16.99 & 18.81 & 83.23 & 15.97 & 16.10 & 70.45 & 17.16 & 19.59 \\
        & PromptCD & \textbf{86.19} & \textbf{3.99} & \textbf{4.43} & \textbf{84.59} & \textbf{13.97} & \textbf{14.17} & \textbf{72.65} & \textbf{13.77} & \textbf{15.93} \\
        \midrule
        \multirow{4}{*}{\textsc{Qwen2.5-7B-instruct}} & \cellcolor{gray!20}Vanilla & \cellcolor{gray!20}73.39 & \cellcolor{gray!20}32.39 & \cellcolor{gray!20}31.27   & \cellcolor{gray!20}43.78 & \cellcolor{gray!20}15.42 & \cellcolor{gray!20}26.05   & \cellcolor{gray!20}75.84 & \cellcolor{gray!20}24.15 & \cellcolor{gray!20}24.15   \\
        & Attr & 82.39 & 17.59 & 17.59 & 90.01 & 33.13 & 26.90 & 75.44 & 20.75 & 21.57 \\
        & Opin & 82.59 & 21.79 & 20.88 & 89.02 & 33.93 & 27.60 & 76.24 & 21.75 & 22.20 \\
        & PromptCD & \textbf{90.39} & \textbf{14.39} & \textbf{13.74} & \textbf{91.82} & \textbf{29.54} & \textbf{24.34} & \textbf{77.64} & \textbf{18.15} & \textbf{18.95} \\
        \bottomrule
    \end{tabular}%
    }
    \label{tab:results_help}
    \vspace{-4mm}
\end{table*}

\begin{table}[t!]
    \centering
    \caption{Performance (\%) of PromptCD on LLM honesty evaluation.}
    \vspace{-2mm}
    \renewcommand{\arraystretch}{1.2}
    \resizebox{\linewidth}{!}{
    \begin{tabular}{l|l|c c c|c c}
        \toprule
        \multirow{2}{*}{\textbf{Model}} & \multirow{2}{*}{\textbf{Method}} 
        & \multicolumn{3}{c|}{\textbf{TruthfulQA(MC)}} 
        & \multicolumn{2}{c}{\textbf{FActScore}} \\
        \cmidrule(lr){3-5} \cmidrule(lr){6-7}
        & & \text{MC1} & \text{MC2} & \text{MC3}
        & \text{Score} & \text{\#facts} \\
        \midrule
         & \cellcolor{gray!20}Vanilla & \cellcolor{gray!20}33.66 & \cellcolor{gray!20}51.29 & \cellcolor{gray!20}24.91 & \cellcolor{gray!20}33.20 & \cellcolor{gray!20}25.60 \\
         \multirow{2}{*}{\textsc{LLaMA2-}} & DoLa & 32.68 & 61.12 & 29.86  & 35.60 & \textbf{54.10}  \\
         \multirow{2}{*}{\textsc{7B-chat}}& SLED & 37.08 & 63.83 &32.92  & 36.00 & 32.40  \\
        & RECITE & 32.19 & 50.38 & 25.06 & 36.10 & 32.40 \\
        & PromptCD & \textbf{42.59} & \textbf{64.05} &\textbf{33.89}  & \textbf{36.60} & 46.40 \\
        \midrule
         & \cellcolor{gray!20}Vanilla & \cellcolor{gray!20}31.95 & \cellcolor{gray!20}48.83 & \cellcolor{gray!20}23.92 & \cellcolor{gray!20}34.20 & \cellcolor{gray!20}20.80 \\
        \multirow{2}{*}{\textsc{LLaMA3-}} &  DoLA & 39.16 & 68.27 & 35.69  & 41.80 & 54.30 \\
        \multirow{2}{*}{\textsc{8B-instruct}} & SLED & 41.37 & 69.17 & 37.88 &30.40 & 14.60  \\
        & RECITE & 41.98 & 61.32 & 33.11 &37.30 & 28.60 \\
        & PromptCD & \textbf{49.21} & \textbf{69.59} & \textbf{43.83} &\textbf{44.50} &\textbf{63.40} \\
        \midrule
         & \cellcolor{gray!20}Vanilla & \cellcolor{gray!20}48.59 & \cellcolor{gray!20}66.24 & \cellcolor{gray!20}37.45  & \cellcolor{gray!20}44.50 & \cellcolor{gray!20}22.5  \\
        \multirow{2}{*}{\textsc{Mistralv0.3-}} & DoLA & 41.12 & 67.99 & 37.51 &46.20 & \textbf{40.5}  \\
        \multirow{2}{*}{\textsc{7B-instruct}} & SLED & 45.41 & 68.43 & 40.35 & 46.30 & 30.5  \\
        & RECITE & 49.79 & 68.89 & 39.40 & 46.30 & 30.5 \\
        & PromptCD & \textbf{50.79} & \textbf{69.58} & \textbf{44.24} & \textbf{47.30} & 26.7 \\
        \midrule
         & \cellcolor{gray!20}Vanilla & \cellcolor{gray!20}46.88 & \cellcolor{gray!20}65.15 & \cellcolor{gray!20}35.22  & \cellcolor{gray!20}29.60 & \cellcolor{gray!20}26.90 \\
        \multirow{2}{*}{\textsc{Qwen2.5-}} & DoLA & 35.74 & 58.76 & 30.68  & 26.20 & 38.10  \\
        \multirow{2}{*}{\textsc{7B-instruct}} & SLED & 44.92 &70.35 & 39.82 & 33.20 & 27.2 \\
        & RECITE & 47.24 &67.61 & 36.90 & 30.80 & \textbf{44.30} \\
        & PromptCD & \textbf{54.95} &\textbf{71.13} & \textbf{45.93} & \textbf{34.30} & 29.00 \\
        \bottomrule
    \end{tabular}
    }
    \label{tab:results_honesty}
\end{table}

\subsubsection{Honesty}

We consider both multiple-choice and open-ended generation tasks to evaluate LLM honesty.
For the multiple-choice setting, we use TruthfulQA~\cite{lin2021truthfulqa} to assess the model’s factual accuracy in short-answer scenarios.
For open-ended generation, we adopt FActScore~\cite{min2023factscore}, which is rated by GPT-4.1-nano.
The remaining settings and configurations are consistent with those in the FActScore paper.
We compare them with four baselines: 1) original decoding, 2) DoLa~\cite{chuang2023dola}, decoding by contrasting layers, 3) SLED~\cite{zhang2024sled}, monitoring self-logit evolution to penalize unstable generations. 4) RECITE~\cite{sun2022recitation}, recitation-based factual generation and our PromptCD.
In the experimental setup for DoLA, we adhered to the same configuration as in the original authors' work, employing the DoLa-static mode throughout. 

\begin{table}[t]
    \centering
    \caption{Performance (\%) of PromptCD on LLM harmlessness evaluation.}
    \vspace{-2mm}
    \renewcommand{\arraystretch}{1.3}
    \resizebox{\linewidth}{!}{
    \begin{tabular}{l|l|c|cccc|c}
        \toprule
        \multirow{2}{*}{\textbf{Method}} & \multirow{2}{*}{\textbf{Mode}} & \textbf{Defense} & \multicolumn{4}{c|}{\textbf{Defense Generalization}} & \multirow{2}{*}{\textbf{Fluency}} \\
        \cmidrule(lr){4-7}
        & & \textbf{Success} & \textbf{onlyQ} & \textbf{otherA} & \textbf{otherQ} & \textbf{otherAQ} & \\
        \midrule
         & \cellcolor{gray!20}Vanilla & \cellcolor{gray!20}40.59 & \cellcolor{gray!20}76.96 & \cellcolor{gray!20}22.41 & \cellcolor{gray!20}41.33 & \cellcolor{gray!20}21.78 & \cellcolor{gray!20}7.01 \\
        {\textsc{LLaMA2-}} & FT-L & 42.52 & 76.22 & 28.37 & 42.89 & 28.85 & 5.94 \\
        {\textsc{7B-chat}} & DINM & 90.00 & 89.93 & 59.26 & 87.41 & 57.63 & 6.61 \\
        & PromptCD & \textbf{96.67} & \textbf{98.67} & \textbf{82.30} & \textbf{96.59} & \textbf{82.37} & \textbf{7.29} \\
        \midrule
         & \cellcolor{gray!20}Vanilla & \cellcolor{gray!20}12.15 & \cellcolor{gray!20}54.81 & \cellcolor{gray!20}18.07 & \cellcolor{gray!20}12.07 & \cellcolor{gray!20}18.74 & \cellcolor{gray!20}\textbf{7.46} \\
        {\textsc{LLaMA3-}} & FT-L & 11.70 & 54.96 & 30.59 & 11.78 & 30.52 & 7.16 \\
        {\textsc{8B-instruct}} & DINM & \textbf{99.26} & \textbf{99.56} & \textbf{99.56} & \textbf{99.11} & \textbf{99.85} & 1.50 \\
        & PromptCD & 95.26 & 92.96 & 84.81 & 95.78 & 83.85 & 5.95 \\
        \midrule
         & \cellcolor{gray!20}Vanilla & \cellcolor{gray!20}9.85 & \cellcolor{gray!20}34.44 & \cellcolor{gray!20}9.04 & \cellcolor{gray!20}9.78 & \cellcolor{gray!20}8.44 & \cellcolor{gray!20}\textbf{7.66} \\
        {\textsc{Mistralv0.3}} & FT-L & 10.67 & 36.67 & 19.00 & 10.22 & 18.44 & 6.65 \\
        {\textsc{7B-instruct}} & DINM & 35.78 & 73.48 & 26.30 & 33.78 & 24.89 & 6.82 \\
        & PromptCD & \textbf{38.67} & \textbf{86.67} & \textbf{36.44} & \textbf{39.78} & \textbf{37.04} & 7.14 \\
        \midrule
         & \cellcolor{gray!20}Vanilla & \cellcolor{gray!20}17.63 & \cellcolor{gray!20}72.59 & \cellcolor{gray!20}7.11 & \cellcolor{gray!20}10.44 & \cellcolor{gray!20}7.04 & \cellcolor{gray!20}\textbf{8.01} \\
        {\textsc{Qwen2.5}} & FT-L & 12.30 & 74.00 & 24.59 & 11.26 & 24.89 & 7.69 \\
        {\textsc{7B-instruct}} & DINM & 26.74 & 84.22 & 34.59 & 24.52 & 33.78 & 7.78 \\
        & PromptCD & \textbf{76.00} & \textbf{88.89} & \textbf{76.89} & \textbf{77.78} & \textbf{76.96} & 6.44 \\
        \bottomrule
    \end{tabular}
    }
    \label{tab:results_harmless}
\end{table}

\subsubsection{Harmlessness}

Harmlessness of LLMs requires the model to refuse generating harmful content when confronted with adversarial inputs.
To this end, we conduct experiments on the SafeEdit benchmark~\cite{wang2024detoxifying}, covering 9 unsafe categories including offensiveness, bias, physical harm, mental harm, illegal activities, ethics violations, privacy breaches, pornography, and political sensitivity.
The benchmark comprises 1,350 test samples, each containing adversarial prompts and corresponding generalization test inputs.
We evaluate Defense Success (DS) for the adversarial inputs in the test set, and Defense Generalization (DG) across multiple scenarios: DG$_{\text{onlyQ}}$ for harmful questions alone, DG$_{\text{otherA}}$ for different attack prompts, DG$_{\text{otherQ}}$ for different harmful questions, and DG$_{\text{otherAQ}}$ for different combinations of both. 
To assess the safety of generated responses, we employ the safety classifier from~\cite{wang2024detoxifying}, which was trained on RoBERTa-large to classify LLM responses as either safe or unsafe.
We also measure fluency via n-gram to ensure response naturalness.
We compare our PromptCD with several baselines: 1) Vanilla, the original aligned model, 2) FT-L, fine-tuning a single layer via causal analysis, 3) DINM~\cite{wang2024detoxifying}, modifying toxic layers via semantic analysis with a single instance.

\subsection{Setup in Vision-Language Models}

We evaluate our method on four benchmark VQA datasets: A-OKVQA~\citep{schwenk2022okvqa}, POPE~\citep{li2023pope}, V$^*$~\citep{v-star}, and TextVQA~\citep{textvqa}.
These datasets cover a diverse spectrum of capabilities, including visual reasoning, perception, and knowledge acquisition.
Notably, for TextVQA, we assess the models' intrinsic visual text recognition skills by relying solely on image-question pairs without external OCR augmentation.
Our experiments employ four VLMs: \textsc{Qwen2.5-VL-Instruct} (3B and 7B)~\citep{qwen2025qwen25vl} and \textsc{LLaVA-1.5} (7B and 13B)~\citep{liu2023visual}.
Regarding input resolution, the Qwen series processes images at $448 \times 448$, while the LLaVA-1.5 series operates at $336 \times 336$.
Greedy decoding is applied across all experiments.

\section{Evaluation Results}

\subsection{Evaluation in Text-Only Models}

\begin{figure}[t]
    \centering
    \includegraphics[width=\linewidth]{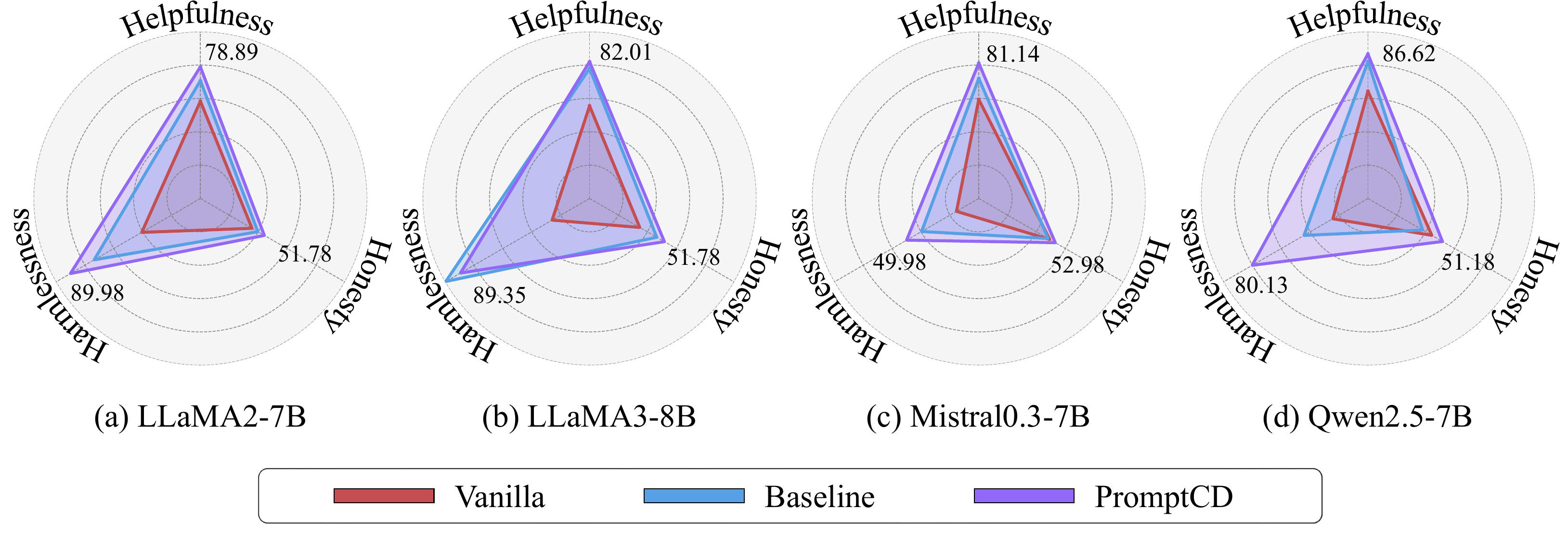}
    \vspace{-6mm}
    \caption{Average performance across the three “3H” dimensions on different models. PromptCD consistently yields substantial improvements over the original models, while the baselines (Opin, DoLa, and DINM) show relatively limited gains.}
    \vspace{-2mm}
    \label{fig:rendar}
\end{figure}

The performance across three distinct behavioral dimensions—helpfulness, honesty, and harmlessness—is systematically reported in Tables~\ref{tab:results_help}, \ref{tab:results_honesty}, and \ref{tab:results_harmless}, respectively.
Additionally, Figure~\ref{fig:rendar} visualizes the average improvements across these ``3H'' dimensions for different models.

\textbf{Helpfulness.}
Our PromptCD demonstrates superior capability in adhering to context instructions, particularly under conflicting scenarios.
It achieves average improvements of 62.64\%, 59.49\%, and 11.08\% in terms of ConR across the three datasets.
Notably, PromptCD consistently maintains the lowest parametric recall (ConR) and memorization ratio (MR).
This underscores its advantage as a decoding-time strategy over prompt-based baselines, as it effectively suppresses the probability of undesired parametric generations.

\textbf{Honesty.}
Decoding strategies, specifically PromptCD and SLED, significantly enhance multiple-choice truthfulness compared to baseline and prompting methods.
PromptCD secures the top performance, achieving the highest MC1 scores on models such as LLaMA2-7B-Chat (42.6\%) and Qwen2.5-7B-Instruct (54.9\%), whereas SLED shows moderate improvements and RECITE provides inconsistent gains.
Furthermore, PromptCD leads in open-ended generation metrics (FActScore), surpassing both the Baseline and DoLa on LLaMA3-8B-Instruct (44.5\%) and Mistral-7B-Instruct (47.3\%).
While DoLa tends to generate more atomic facts, it often sacrifices precision. Conversely, SLED maintains strong factual consistency, and RECITE offers slight improvements in precision.

\textbf{Harmlessness.}
PromptCD exhibits robust safety alignment across various models while preserving response fluency—a trade-off often mishandled by other knowledge editing methods.
On LLaMA2-7B-Chat, PromptCD achieves the best results with a Defense Success (DS) of 96.67\% and an average DG performance of 82.37\%, substantially outperforming the vanilla baseline (40.59\% DS) while maintaining excellent fluency (7.29).
Similarly, on Mistral-7B-Instruct-v0.3, PromptCD shows superior performance with a DS of 38.67\% compared to the vanilla 9.85\%.
Although DINM achieves higher defense success rates on Meta-Llama-3-8B-Instruct, it does so at the cost of substantially degraded fluency (1.50 vs. 5.95).
Overall, PromptCD effectively balances harmlessness with generation quality, positioning it as a practical solution for detoxifying LLMs without compromising naturalness.

In summary, PromptCD achieves consistently superior performance across all three key behavioral dimensions of LLMs.
Unlike existing methods that are restricted to a single behavioral objective, PromptCD offers the flexibility to switch target behaviors simply by adjusting the polarity prompt, thereby enhancing both the effectiveness and adaptability of test-time intervention.

\begin{table}[t]
    \centering
    \renewcommand{\arraystretch}{1.3}
    \caption{Performance (\%) of PromptCD on VLMs across different layer intervals.}
    \vspace{-2mm}
    \renewcommand{\arraystretch}{1}
    \resizebox{\linewidth}{!}{
    \setlength{\tabcolsep}{6pt}
    \begin{tabular}{l|l|cccc}
        \toprule
        \textbf{Model} & \textbf{Layers} $\mathcal{L}$ & \textbf{A-OKVQA} & \textbf{POPE} & \textbf{V$^*$} & \textbf{TextVQA} \\
        \midrule 
        \multirow{4}{*}{\textsc{Qwen2.5-VL-3B}} 
        & \cellcolor{gray!20}Vanilla & \cellcolor{gray!20}73.0 & \cellcolor{gray!20}86.9 & \cellcolor{gray!20}50.3 & \cellcolor{gray!20}72.8 \\
        & [10, 15] & 74.0 & 86.9 & 53.4 & 73.0 \\
        & [15, 20] & 76.8 & 87.7 & 56.0 & 76.0 \\
        & [20, 25] & \textbf{78.3} & \textbf{87.9} & \textbf{57.1} & \textbf{76.3} \\
        \midrule
        \multirow{4}{*}{\textsc{Qwen2.5-VL-7B}} 
        & \cellcolor{gray!20}Vanilla & \cellcolor{gray!20}75.0 & \cellcolor{gray!20}87.0 & \cellcolor{gray!20}50.8 & \cellcolor{gray!20}75.0 \\
        & [10, 15] & 75.0 & 87.0 & 51.3 & 75.0 \\
        & [15, 20] & 77.1 & 88.4 & 57.6 & 79.5 \\
        & [20, 25] & \textbf{78.0} & \textbf{88.6} & \textbf{58.1} & \textbf{81.7} \\
        \midrule
        \multirow{4}{*}{\textsc{LLaVA-1.5-7B}} 
        & \cellcolor{gray!20}Vanilla & \cellcolor{gray!20}71.5 & \cellcolor{gray!20}83.6 & \cellcolor{gray!20}38.7 & \cellcolor{gray!20}47.8 \\
        & [10, 15] & 71.5 & 84.5 & 48.2 & 49.2 \\
        & [15, 20] & 74.2 & 87.5 & 65.4 & 56.4 \\
        & [20, 25] & \textbf{75.4} & \textbf{89.0} & \textbf{66.5} & \textbf{58.2} \\
        \midrule
        \multirow{4}{*}{\textsc{LLaVA-1.5-13B}} 
        & \cellcolor{gray!20}Vanilla & \cellcolor{gray!20}75.7 & \cellcolor{gray!20}84.6 & \cellcolor{gray!20}42.4 & \cellcolor{gray!20}57.1 \\
        & [10, 15] & 75.7 & 85.0 & 52.9 & 57.4 \\
        & [15, 20] & 76.8 & 88.6 & 69.1 & 59.4 \\
        & [20, 25] & \textbf{76.9} & \textbf{90.1} & \textbf{70.0} & \textbf{61.2} \\
        \bottomrule
    \end{tabular}%
    }
    \label{tab:vlm_results}
\end{table}

\begin{table}[t]
    \centering
    \caption{Ablation results of the adjustment coefficient $\gamma$ on \textsc{LLaMA2-7B-chat} models across the `3H' dimensions. Performance is reported using the average values of ConR, MC-AVG, and DG-AVG, respectively.}
    \renewcommand{\arraystretch}{1.2}
\resizebox{\linewidth}{!}{
    \begin{tabular}{lccc}
        \toprule
        \textbf{Model} & \textbf{$\gamma=0.2$} &  \textbf{$\gamma=0.5$} &  \textbf{$\gamma=0.8$} \\
        \midrule
        \textbf{Helpfulness} (ConR) & 77.32 & \bf 78.89 & 70.53 \\
        \textbf{Honesty} (MC-AVG) & 39.59  & \bf 46.84 & 44.87 \\
        \textbf{Harmlessness} (DG-AVG) & 83.29  & \bf 89.98 & 85.34 \\
        \bottomrule
    \end{tabular}
    }
    \label{tab:coefficient}
\end{table}

\subsection{Evaluation in Vision-Language Models}
\label{sec:vlm_results}

\begin{figure}[t]
    \centering
    \includegraphics[width=\linewidth]{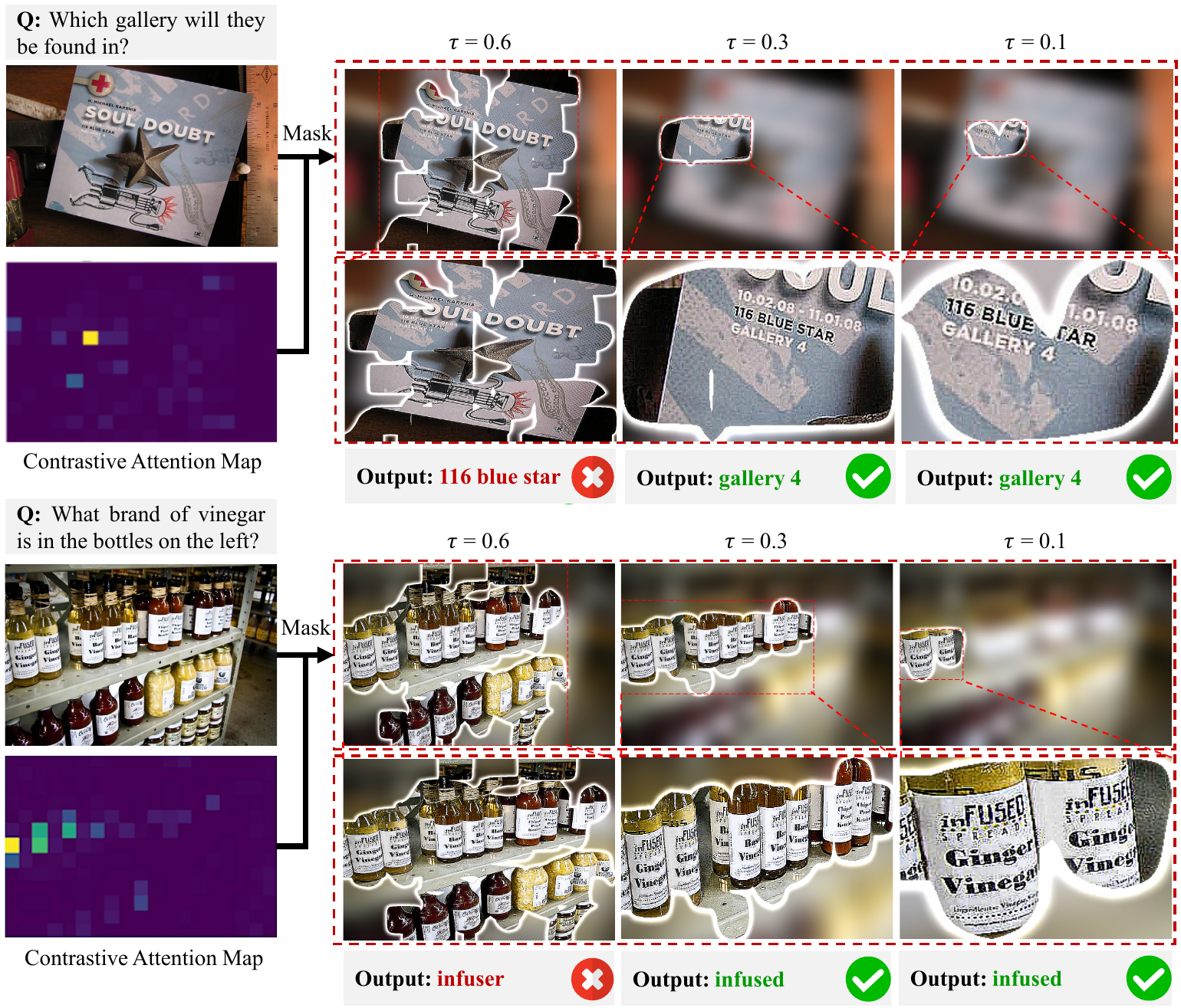}
    \vspace{-6mm}
    \caption{Visualization of contrastive attention refinement in VQA. 
    As the threshold $\tau$ decreases from 0.6 to 0.1, PromptCD progressively suppresses visual noise to isolate request-relevant regions. 
    Our PromptCD precise grounding corrects initial hallucinations (Red $\times$) and yields accurate answers (Green $\checkmark$).}
    \label{fig:vqa_vis}
    \vspace{-4mm}
\end{figure}

Table~\ref{tab:vlm_results} demonstrates PromptCD's consistent performance enhancement across all evaluated VLMs and VQA datasets through cross-modal contrastive attention refinement.
Earlier-generation models exhibit substantially greater improvements than recent counterparts: \textsc{LLaVA-1.5-7B} achieves 71.83\% relative improvement on V$^*$ versus 14.37\% for \textsc{Qwen2.5-VL-7B}, and 21.76\% gain on TextVQA (47.8\% $\to$ 58.2\%) versus 8.93\% for \textsc{Qwen2.5-VL-7B} (75.0\% $\to$ 81.7\%).
Performance gains vary significantly across datasets: V$^*$ shows the most substantial improvements, with \textsc{LLaVA-1.5-13B} achieving 70.0\% versus 42.4\% baseline (65.09\% relative improvement), while POPE exhibits moderate gains (90.1\% from 84.6\%, 6.50\% improvement).
This suggests attention refinement particularly benefits tasks requiring precise spatial localization and fine-grained visual grounding.
Models with limited visual reasoning capabilities suffer more from attention dispersion over task-irrelevant regions, thereby benefiting more from PromptCD's contrastive attention-guided focusing mechanisms.
As shown in Figure~\ref{fig:vqa_vis}, PromptCD effectively generates high-fidelity attention masks that filter background noise, shifting focus from broad, distracted regions to precise, request-relevant visual cues as threshold $\tau$ tightens, thereby correcting hallucinations.

\subsection{Ablation Study}
We investigate the sensitivity of PromptCD to the adjustment coefficient $\gamma$, which governs the magnitude of the penalty applied to the negative logits (i.e., the strength of suppression).
For all reported experiments on text-only LLMs, we uniformly set $\gamma$ to 0.5.
The detailed ablation results are presented in Table~\ref{tab:coefficient}.
We observe that performance follows an inverted U-shape trend: an excessively small $\gamma$ fails to sufficiently amplify the contrast between positive and negative contexts, while an overly large $\gamma$ may disrupt the model's linguistic coherence.
The results empirically confirm that a moderate $\gamma$ yields the optimal balance, effectively steering the behavior without degrading generation quality.

For vision tasks, as shown in Table~\ref{tab:vlm_results}, layer-wise analysis reveals a consistent performance ordering across all VLMs: [20, 25], [15, 20], and [10, 15]. Taking \textsc{LLaVA-1.5-7B} on TextVQA as an example, deep layers [20, 25] achieve 21.76\% improvement, middle layers [15, 20] reach 17.99\%, while early layers [10, 15] attain only 2.93\%. This hierarchical pattern aligns with established attention mechanisms: early layers perform global visual scanning with high entropy, while deep layers progressively focus on task-relevant semantic patterns, making them optimal for extracting request-consistent visual attention through polarity-prompt contrastive decoding. The optimal configuration [20, 25] consistently outperforms other intervals with absolute accuracy improvements ranging from 1.2\% to 27.8\%, validating the effectiveness of deep-layer attention extraction.

\begin{table*}[t!]
    \centering
    \caption{Performance comparison of PromptCD against external tool-based approaches and ViCrop on TextVQA dataset: \\ accuracy (\%) and inference time (seconds) overhead per sample.}
    \vspace{-2mm}
    \resizebox{0.8\linewidth}{!}{
    \begin{tabular}{l|c|ccc|ccc|c}
    \toprule
    \multirow{2}{*}{\textbf{Method}} & \multirow{2}{*}{\textbf{Original}} & \multirow{2}{*}{\textbf{SAM}} & \multirow{2}{*}{\textbf{YOLO}} & \multirow{2}{*}{\textbf{CLIP}} & \multicolumn{3}{c|}{\textbf{ViCrop}} & \multirow{2}{*}{\textbf{PromptCD}} \\
    \cmidrule(lr){6-8}
    & & & & & \texttt{rel-att} & \texttt{grad-att} & \texttt{pure-grad} & \\
    \midrule
    Accuracy & 47.80 & 49.42 & 48.84 & 48.55 & 55.17 & 56.06 & 51.67 & \goodmetric{58.2} \\
    \midrule
    GPU Time & 0.17 & 3.33 & 0.35 & 1.07 & 1.16 & 0.89 & 2.36 & 1.34 \\
    \bottomrule
    \end{tabular}
    }
    \vspace{-2mm}
    \label{tab:vis_time}
\end{table*}

\subsection{Inference Latency Analysis}
Since PromptCD necessitates dual forward passes at each decoding step, it inevitably incurs computational overhead.
Table~\ref{tab:Latency} reports the inference latency comparison evaluated on the NQ dataset, indicating that our method results in a $1.6\times$--$1.8\times$ overhead in latency and throughput compared to standard decoding.
While this cost is non-negligible, we argue that it is a justifiable trade-off for safety-critical and high-stakes applications—such as those in medical, financial, or legal domains.
In these scenarios, factual errors, unmanaged knowledge conflicts, or safety violations can have severe consequences.
Therefore, the marginal increase in computational cost is outweighed by the substantial gains in reliability, trustworthiness, and the capability for precise, human-in-the-loop behavioral governance.

\begin{table}[t]
  \centering
  \caption{Comparison of inference latency and throughput for PromptCD.}
  \vspace{-2mm}
  \label{tab:Latency}
  \renewcommand{\arraystretch}{1.2}
\resizebox{\linewidth}{!}{
  \begin{tabular}{llcc}
    \toprule
    \textbf{Model} & \textbf{Method} & \textbf{Latency (ms/token)} & \textbf{Throughput (token/s)} \\
    \midrule
    \multirow{2}{*}{\textsc{LLaMA2-7B}}
      & Vanilla   & 36.03 ($\times$1.00) & 27.76 ($\times$1.00) \\
      & PromptCD & 64.85 ($\times$1.79) & 16.64 ($\times$0.60) \\
    \midrule
    \multirow{2}{*}{\textsc{LLaMA2-13B}}
      & Vanilla   & 51.41 ($\times$1.00) & 19.45 ($\times$1.00) \\
      & PromptCD & 89.45 ($\times$1.74) & 12.25 ($\times$0.63) \\
    \midrule
    \multirow{2}{*}{\textsc{LLaMA3-8B}}
      & Vanilla   & 42.17 ($\times$1.00) & 23.92 ($\times$1.00) \\
      & PromptCD & 69.77 ($\times$1.65) & 14.08 ($\times$0.59) \\
    \bottomrule
  \end{tabular}
}
\vspace{-2mm}
\end{table}

\begin{table}[t!]
\centering
\renewcommand{\arraystretch}{1.0}
\caption{Ablation study on the NQ dataset evaluating the effect of the Adaptive Plausibility Constraint (APC) on generation quality (Hit Rate).}
\vspace{-2mm}
\resizebox{0.9\linewidth}{!}{
\begin{tabular}{lccc}
    \toprule
    \multirow{2}{*}{\textbf{Model}} & \multirow{2}{*}{\textbf{\textit{Vanilla}}} & \multicolumn{2}{c}{\textbf{PromptCD}} \\
    \cmidrule(lr){3-4}
    & & w/ APC & w/o APC \\
    \midrule
    \textsc{LLaMA2-7B} & \bf 78.9 & 77.8 & {30.7} \\
    \textsc{LLaMA3-8B} & 83.7 & \bf 85.5 & {53.8} \\
    \textsc{Mistralv0.3-7B} & \bf 92.4 & 91.7 & {49.9} \\
    \textsc{Qwen2.5-7B} & 88.8 & \bf 89.8 & {62.4} \\
    \bottomrule
\end{tabular}
}
\label{tab:apc_hitrate}
\vspace{-2mm}
\end{table}

For vision-language tasks, Table~\ref{tab:vis_time} compares PromptCD against external tools such as SAM~\citep{kirillov2023segment}, YOLO~\citep{redmon2016you}, CLIP~\citep{radford2021learning} and ViCrop~\citep{zhang2025mllms}.
PromptCD achieves the highest accuracy with competitive inference times, establishing a favorable balance between visual grounding precision and computational cost.

\subsection{Robustness against Cross-Task Interference}
A critical concern in alignment interventions is the "alignment tax"—specifically, whether enhancing a target behavior comes at the cost of degrading the model's general generation quality.
To investigate this, we conduct an analysis using the {NQ dataset} under the Helpfulness setting.
We define a metric termed {Hit Rate}, which measures whether the model's output contains either the original parametric answer or the correct gold answer provided in the context.
This metric effectively serves as a proxy for linguistic coherence and meaningfulness; a drop in Hit Rate implies the model is generating irrelevant or hallucinated content (gibberish) rather than valid answers.

The results are summarized in Table~\ref{tab:apc_hitrate}.
We observe that without Adaptive
Plausibility Constraint (APC), the Hit Rate drops significantly.
This suggests that unconstrained logit adjustments can severely disturb the model's distribution, causing it to deviate from both its internal knowledge and the external context.
In contrast, when equipped with APC, the performance remains comparable to the Vanilla baseline.
This finding demonstrates that APC effectively acts as a safety filter, restricting PromptCD's intervention to plausible tokens and preserving the model's fundamental ability to generate coherent and relevant responses.

\section{Applications and Interpretability}


\begin{table}[t!]
\centering
\small
\renewcommand{\arraystretch}{1.0}
\caption{Experimental results (accuracy \%) using \textsc{LLaMA2-chat} models. We conduct experiments with full-batch edit memory to evaluate the performance of memory-based Knowledge Editing.}
\vspace{-2mm}
\resizebox{\linewidth}{!}{
\begin{tabular}{llccc}
\toprule
\multirow{2}{*}{\textbf{Model}} & \multirow{2}{*}{\textbf{Method}} &  \multicolumn{3}{c}{\textbf{\textsc{MQuAKE}}} \\ \cmidrule(lr){3-5} & & {\textbf{\textsc{3k}}} &{\textbf{\textsc{2002}}} & {\textbf{\textsc{hard}}}\\
\midrule
   \multirow{4}{*}{\textsc{LLaMA2-7B-chat}} & IKE  & 20.7 & \bf 20.6 & 2.3\\
    & \ + PromptCD & \bf 22.4 & 20.4 & \bf 3.8\\
   \cmidrule(lr){2-5}
    & MeLLo& 32.6 & 40.8 & 5.1\\
   & \ + PromptCD & \bf 43.1 & \bf 45.8 & \bf 5.8\\
\midrule
  \multirow{4}{*}{\textsc{LLaMA2-13B-chat}} & IKE  & 19.4 & \bf 18.8 & 2.7\\
   & \ + PromptCD & \bf 20.6 & 18.4 & \bf 3.5\\
  \cmidrule(lr){2-5}
   & MeLLo & 33.4 & 35.9 & 3.9\\
   & \  + PromptCD & \bf 36.8 & \bf 38.2 & \bf 6.2\\
\bottomrule
\end{tabular}
}
\label{tab:ice_all}
\end{table}

\subsection{Application to Knowledge Editing}
\label{sec:app_ke}

Although LLMs accumulate substantial factual knowledge during pretraining, this parametric information often becomes outdated, thereby compromising model reliability~\citep{chen2023combating, zhang2023sirens, huang2023survey}.
Given the prohibitive cost of retraining, Knowledge Editing (KE)~\citep{sinitsin2020editable, de2021editing, mitchell2022memory, yao2023editing, li2025reinforced} has emerged as a vital technique to update LLMs by injecting new information or modifying existing parameters.
In this section, we apply our method to In-Context Editing (ICE)~\citep{cohen2024evaluating, bi2024adaptive, bi2024struedit}, a non-invasive paradigm widely adopted in Retrieval-Augmented Generation (RAG) systems~\citep{guu2020realmretrievalaugmentedlanguagemodel, izacard2021leveragingpassageretrievalgenerative} that corrects outdated knowledge solely through context without altering model weights.

\begin{figure}[t!]
    \centering
    \includegraphics[width=\linewidth]{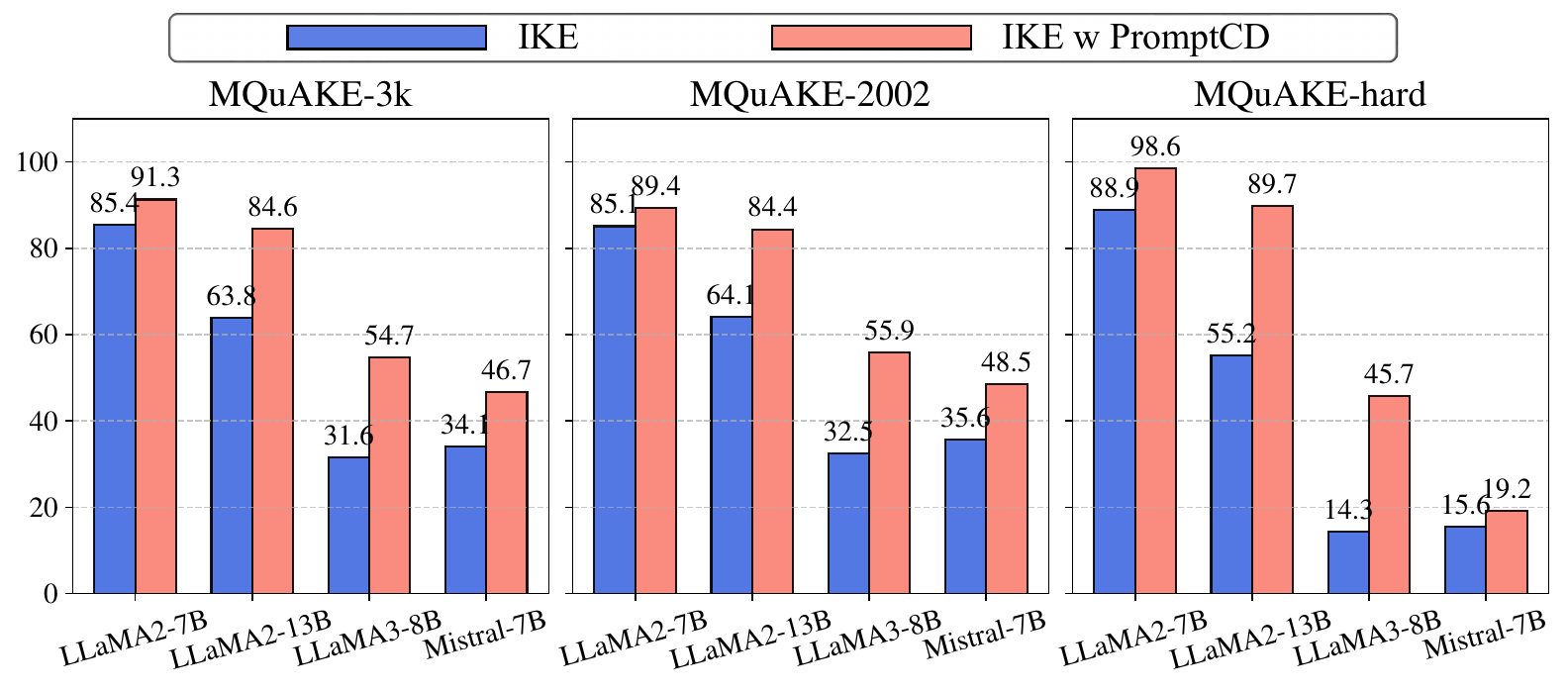}
    \vspace{-6mm}
    \caption{Performance comparison (\%) between standard IKE and IKE enhanced with PromptCD across various models and datasets.  We set the edit memory batch size to 1 to evaluate the fundamental capability of direct knowledge editing.}
    \vspace{-2mm}
    \label{fig:ICE_1}
\end{figure}

\begin{figure}[t]
    \centering
    \includegraphics[width=\linewidth]{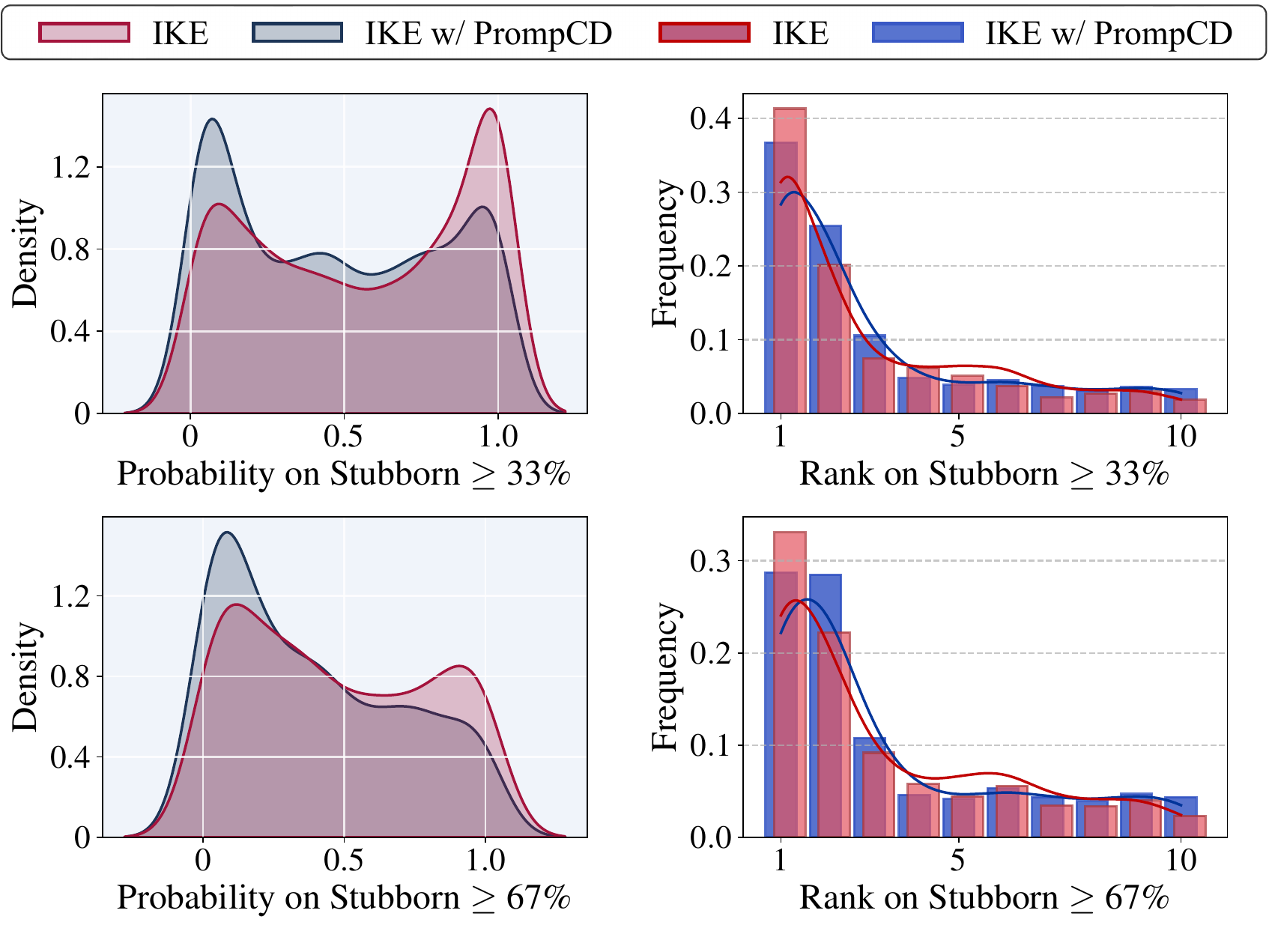}
    \vspace{-6mm}
    \caption{Probability statistics of edited new knowledge on the \textsc{MQuAKE-Stubborn} dataset using \llamaa. The probabilities are computed by applying softmax to the model’s token logits.}
    \vspace{-2mm}
    \label{fig:log_rank}
\end{figure}

To assess the practical utility of PromptCD in realistic application scenarios, we extend our evaluation to the task of ICE, integrating our method into two representative baselines:
\textbf{IKE}~\citep{cohen2024evaluating}, which guides LLMs to edit knowledge via contextual demonstrations; and
\textbf{MeLLo}~\citep{zhong2023mquake}, which handles multi-hop editing by decomposing complex questions and retrieving conflicting information iteratively.
Our experiments are conducted on \textsc{MQuAKE-3k}~\citep{zhong2023mquake} and the more challenging \textsc{MQuAKE-2002/hard}~\citep{wang2024deepedit}.

Figure~\ref{fig:ICE_1} illustrates the performance of IKE with a memory batch size of 1.
We observe a counter-intuitive phenomenon: as model size or training strength increases, the model's parametric knowledge becomes more "stubborn," leading to a decline in standard ICE accuracy.
This suggests that stronger models suffer more from the conflict between intrinsic priors and external context.
In contrast, IKE enhanced with PromptCD consistently outperforms the vanilla version across all model scales, indicating that our approach effectively mitigates the resistance of stubborn knowledge, allowing even strong models to adapt to new information.

Table~\ref{tab:ice_all} presents the results of full-batch experiments (1,000 instances), simulating a realistic RAG retrieval environment.
PromptCD yields consistent improvements for both IKE and MeLLo.
These results underscore the potential of PromptCD for modern RAG systems, demonstrating its ability to enhance both faithfulness and controllability without requiring architectural modifications.

\subsection{Metamorphosis of Stubborn Knowledge}
\label{sec:metamorphosis}

To understand why PromptCD succeeds where standard methods fail, we revisit the critical challenge identified in Section~\ref{sec:explore}: the misalignment between Latent Alignment and Behavioral Emergence.
As defined previously, "stubborn knowledge" represents a failure state where the inertia of parametric priors suppresses the realization of new knowledge.
We investigate the internal dynamics of this phenomenon using the \textsc{Stubborn} dataset (derived from \textsc{MQuAKE-3k}) and apply the Knowledge Token Capturing Algorithm (Algorithm~\ref{alg:alg}) from Section~\ref{sec:token_analysis} to trace the rank and probability mass of the first distinctive faithful token.

Figure~\ref{fig:log_rank} visualizes the distribution of these key tokens, revealing a distinct "Metamorphosis" of the probability landscape.
Under the standard ICE setting, consistent with our earlier findings, the faithful tokens often languish in the "long tail" or hover just below the decision threshold.
This empirical evidence reconfirms that standard prompting functions merely as a weak probabilistic bias, often insufficient to dislodge confident parametric priors.
In sharp contrast, the application of PromptCD induces a radical shift in the distribution toward high-probability regions.
By actively modulating the logits based on the contrast between positive and negative signals, our method effectively "pushes" these latent faithful tokens across the decision boundary.

This intervention significantly increases the likelihood of decoding context-faithful tokens at critical positions, resulting in a marked rise in the frequency of high-ranking generations.
The analysis confirms that PromptCD functions exactly as the decisive control mechanism hypothesized in our motivation: it actively amplifies the latent tendency of the model, transforming tentative internal signals into stable, manifest behavioral realizations.

\section{Related Work}

\subsection{Behavioral Alignment of LLMs}

The alignment of LLMs with human preferences and values has emerged as a central research focus~\citep{gabriel2020artificial,ji2023ai}.
Dominant approaches primarily rely on Supervised Fine-Tuning (SFT) and Reinforcement Learning from Human Feedback (RLHF), alongside recent variants like Direct Preference Optimization (DPO)~\citep{ouyang2022training,christiano2017deep,rafailov2023direct}.
While these methods have proven effective in improving helpfulness and safety~\citep{bai2022constitutional}, and multi-objective frameworks attempt to address trade-offs among honesty and harmlessness~\citep{askell2021general,yuan2025autodrive}, they suffer from inherent rigidity and high computational costs~\citep{casper2023open,hendrycks2020aligning}.
Specifically, once aligned, the model's behavioral patterns are frozen, making it difficult to adaptively switch behaviors for diverse scenarios without expensive retraining.
Consequently, recent research has shifted towards lightweight, training-free alternatives~\citep{li2023self,park2023generative}.
Strategies such as self-critique, self-consistency, and uncertainty-based signals have been proposed to provide weak supervision or refinement~\citep{perez2023discovering,madaan2023self}.
Furthermore, large-scale audits continue to reveal persistent data and value biases~\citep{santurkar2023whose,bang2024measuring}, underscoring the necessity for more flexible, inference-stage intervention mechanisms.

\subsection{Test-Time Behavior Enhancement}

Beyond these training-free strategies, a complementary line of work focuses on test-time interventions that adjust model behaviour dynamically during inference~\citep{lin2022teaching,li2022contrastive,bi2025parameters}. Such methods operate by modulating inputs, representations, or decoding dynamics without retraining. Prompt-based steering augments inputs with behavioural cues or exemplars to guide responses~\citep{fernando2023promptbreeder,pryzant2023automatic}, whereas activation editing directly manipulates internal states to induce desired traits~\citep{kong2024aligning,turner2023activation}. More recent decoding-time alignment methods alter token probabilities or logits during generation~\citep{chuang2023dola,liu2024decoding}. However, these approaches are typically tailored to fixed objectives (e.g., factuality or safety) and lack a unified mechanism for flexible behavioural steering. Building on this insight, our method introduces polarity prompts to enable general, low-cost alignment control through contrastive guidance at decoding time.

\subsection{Contrastive Mechanisms in Generation}

Contrastive Decoding (CD) has emerged as a powerful paradigm to enhance generation quality by amplifying the difference between a reliable distribution and a defective one. Existing CD methods primarily vary in their source of contrast: Model-level contrast (e.g., CD~\citep{li2022contrastive}) contrasts expert models with amateur models to improve fluency. Layer-level contrast (e.g., DoLa~\citep{chuang2023dola}, SLED~\citep{zhang2024sled}) contrasts mature final layers with premature early layers to isolate factual knowledge. Context-level contrast (e.g., ICD~\citep{zhang2023alleviating}) contrasts standard context with hallucinated context to suppress errors. Other approaches like TruthX~\citep{zhang2024truthx} and HACL~\citep{wang2023llm} apply contrastive principles within the representation space during training or inference to mitigate hallucinations. 
PAC\citep{zhai-etal-2025-parameter} traces critical transmission paths across all layers, treating less important pathways as negatives, while UniKE~\citep{pan2024towards} extends this idea to multimodal LLMs through semantic truthfulness space disentanglement.
Different from these works, PromptCD introduces Instruction-level contrast. By constructing explicit Positive and Negative prompts, we dynamically synthesize the contrasting distributions required for decoding. This eliminates the need for auxiliary models (as in CD) or layer selection heuristics (as in DoLa), offering a unified framework to steer arbitrary behaviors defined by natural language.

\section{Conclusion}\label{conclusion}

In this work, we propose PromptCD, a polarity-prompt–based contrastive decoding framework for enhancing target behaviors at test time.
PromptCD constructs paired positive and negative guiding prompts for a specified behavioral objective and contrasts the resulting probability distributions during decoding to amplify tokens aligned with the desired behavior.
Extensive experiments on the “3H” alignment dimensions—helpfulness, honesty, and harmlessness—demonstrate that PromptCD effectively strengthens each target behavior and generalizes well across tasks without requiring retraining.
Moreover, its application to knowledge editing further highlights its interpretability and performance advantages.
Overall, this work paves the way toward integrating both behavioral alignment and flexible test-time enhancement in large language models.

\bibliography{new_reference}

@article{wei2022chain,
  title={{Chain-of-thought prompting elicits reasoning in large language models}},
  author={Wei, Jason and Wang, Xuezhi and Schuurmans, Dale and Bosma, Maarten and Xia, Fei and Chi, Ed and Le, Quoc V and Zhou, Denny and others},
  journal={Advances in neural information processing systems},
  volume={35},
  pages={24824--24837},
  year={2022}
}

@article{touvron2023llama,
  title={{Llama: Open and efficient foundation language models}},
  author={Touvron, Hugo and Lavril, Thibaut and Izacard, Gautier and Martinet, Xavier and Lachaux, Marie-Anne and Lacroix, Timoth{\'e}e and Rozi{\`e}re, Baptiste and Goyal, Naman and Hambro, Eric and Azhar, Faisal and others},
  journal={arXiv preprint arXiv:2302.13971},
  year={2023}
}

@article{ouyang2022training,
  title={{Training language models to follow instructions with human feedback}},
  author={Ouyang, Long and Wu, Jeffrey and Jiang, Xu and Almeida, Diogo and Wainwright, Carroll and Mishkin, Pamela and Zhang, Chong and Agarwal, Sandhini and Slama, Katarina and Ray, Alex and others},
  journal={Advances in neural information processing systems},
  volume={35},
  pages={27730--27744},
  year={2022}
}

@article{muennighoff2025s1,
  title={{s1: Simple test-time scaling}},
  author={Muennighoff, Niklas and Yang, Zitong and Shi, Weijia and Li, Xiang Lisa and Fei-Fei, Li and Hajishirzi, Hannaneh and Zettlemoyer, Luke and Liang, Percy and Cand{\`e}s, Emmanuel and Hashimoto, Tatsunori},
  journal={arXiv preprint arXiv:2501.19393},
  year={2025}
}

@inproceedings{radford2021learning,
  title={{Learning transferable visual models from natural language supervision}},
  author={Radford, Alec and Kim, Jong Wook and Hallacy, Chris and Ramesh, Aditya and Goh, Gabriel and Agarwal, Sandhini and Sastry, Girish and Askell, Amanda and Mishkin, Pamela and Clark, Jack and others},
  booktitle={International conference on machine learning},
  pages={8748--8763},
  year={2021},
  organization={PMLR}
}

@article{bi2025parameters,
  title={Parameters vs. Context: Fine-Grained Control of Knowledge Reliance in Language Models},
  author={Bi, Baolong and Liu, Shenghua and Wang, Yiwei and Xu, Yilong and Fang, Junfeng and Mei, Lingrui and Cheng, Xueqi},
  journal={arXiv preprint arXiv:2503.15888},
  year={2025}
}

@article{bi2024context,
  title={Context-DPO: Aligning Language Models for Context-Faithfulness},
  author={Bi, Baolong and Huang, Shaohan and Wang, Yiwei and Yang, Tianchi and Zhang, Zihan and Huang, Haizhen and Mei, Lingrui and Fang, Junfeng and Li, Zehao and Wei, Furu and others},
  journal={arXiv preprint arXiv:2412.15280},
  year={2024}
}

@article{chuang2023dola,
  title={Dola: Decoding by contrasting layers improves factuality in large language models},
  author={Chuang, Yung-Sung and Xie, Yujia and Luo, Hongyin and Kim, Yoon and Glass, James and He, Pengcheng},
  journal={arXiv preprint arXiv:2309.03883},
  year={2023}
}

@article{li2025reinforced,
  title={Reinforced Lifelong Editing for Language Models},
  author={Li, Zherui and Jiang, Houcheng and Chen, Hao and Bi, Baolong and Zhou, Zhenhong and Sun, Fei and Fang, Junfeng and Wang, Xiang},
  journal={arXiv preprint arXiv:2502.05759},
  year={2025}
}

@article{bi2024struedit,
  title={Struedit: Structured outputs enable the fast and accurate knowledge editing for large language models},
  author={Bi, Baolong and Liu, Shenghua and Wang, Yiwei and Mei, Lingrui and Gao, Hongcheng and Fang, Junfeng and Cheng, Xueqi},
  journal={arXiv preprint arXiv:2409.10132},
  year={2024}
}

@article{bi2024adaptive,
  title={Adaptive token biaser: Knowledge editing via biasing key entities},
  author={Bi, Baolong and Liu, Shenghua and Wang, Yiwei and Mei, Lingrui and Gao, Hongcheng and Xu, Yilong and Cheng, Xueqi},
  journal={arXiv preprint arXiv:2406.12468},
  year={2024}
}

@article{achiam2023gpt,
  title   = {Gpt-4 technical report},
  author  = {Achiam, Josh and Adler, Steven and Agarwal, Sandhini and Ahmad, Lama and Akkaya, Ilge and Aleman, Florencia Leoni and Almeida, Diogo and Altenschmidt, Janko and Altman, Sam and Anadkat, Shyamal and others},
  journal = {arXiv preprint arXiv:2303.08774},
  year    = {2023}
}

@article{yang2024qwen2,
  title   = {Qwen2. 5 technical report},
  author  = {Yang, An and Yang, Baosong and Zhang, Beichen and Hui, Binyuan and Zheng, Bo and Yu, Bowen and Li, Chengyuan and Liu, Dayiheng and Huang, Fei and Wei, Haoran and others},
  journal = {arXiv preprint arXiv:2412.15115},
  year    = {2024}
}

@article{grattafiori2024llama,
  title   = {The llama 3 herd of models},
  author  = {Grattafiori, Aaron and Dubey, Abhimanyu and Jauhri, Abhinav and Pandey, Abhinav and Kadian, Abhishek and Al-Dahle, Ahmad and Letman, Aiesha and Mathur, Akhil and Schelten, Alan and Vaughan, Alex and others},
  journal = {arXiv preprint arXiv:2407.21783},
  year    = {2024}
}

@misc{guo2024controllablepreferenceoptimizationcontrollable,
  title         = {Controllable Preference Optimization: Toward Controllable Multi-Objective Alignment},
  author        = {Yiju Guo and Ganqu Cui and Lifan Yuan and Ning Ding and Zexu Sun and Bowen Sun and Huimin Chen and Ruobing Xie and Jie Zhou and Yankai Lin and Zhiyuan Liu and Maosong Sun},
  year          = {2024},
  eprint        = {2402.19085},
  archiveprefix = {arXiv},
  primaryclass  = {cs.CL},
  url           = {https://arxiv.org/abs/2402.19085}
}

@article{yang2023alignment,
  title   = {Alignment for honesty},
  author  = {Yang, Yuqing and Chern, Ethan and Qiu, Xipeng and Neubig, Graham and Liu, Pengfei},
  journal = {arXiv preprint arXiv:2312.07000},
  year    = {2023}
}

@article{tian2023fine,
  title   = {Fine-tuning language models for factuality},
  author  = {Tian, Katherine and Mitchell, Eric and Yao, Huaxiu and Manning, Christopher D and Finn, Chelsea},
  journal = {arXiv preprint arXiv:2311.08401},
  year    = {2023}
}

@article{snell2024scaling,
  title   = {Scaling llm test-time compute optimally can be more effective than scaling model parameters},
  author  = {Snell, Charlie and Lee, Jaehoon and Xu, Kelvin and Kumar, Aviral},
  journal = {arXiv preprint arXiv:2408.03314},
  year    = {2024}
}

@article{sun2022recitation,
  title   = {Recitation-augmented language models},
  author  = {Sun, Zhiqing and Wang, Xuezhi and Tay, Yi and Yang, Yiming and Zhou, Denny},
  journal = {arXiv preprint arXiv:2210.01296},
  year    = {2022}
}

@article{zhou2023context,
  title   = {Context-faithful prompting for large language models},
  author  = {Zhou, Wenxuan and Zhang, Sheng and Poon, Hoifung and Chen, Muhao},
  journal = {arXiv preprint arXiv:2303.11315},
  year    = {2023}
}

@misc{li2024inferencetimeinterventionelicitingtruthful,
  title         = {Inference-Time Intervention: Eliciting Truthful Answers from a Language Model},
  author        = {Kenneth Li and Oam Patel and Fernanda Viégas and Hanspeter Pfister and Martin Wattenberg},
  year          = {2024},
  eprint        = {2306.03341},
  archiveprefix = {arXiv},
  primaryclass  = {cs.LG},
  url           = {https://arxiv.org/abs/2306.03341}
}

@misc{huang2023survey,
  archiveprefix = {arXiv},
  author        = {Lei Huang and Weijiang Yu and Weitao Ma and Weihong Zhong and Zhangyin Feng and Haotian Wang and Qianglong Chen and Weihua Peng and Xiaocheng Feng and Bing Qin and Ting Liu},
  eprint        = {2311.05232},
  primaryclass  = {cs.CL},
  title         = {A Survey on Hallucination in Large Language Models: Principles, Taxonomy, Challenges, and Open Questions},
  year          = {2023}
}

@misc{zhang2023sirens,
  archiveprefix = {arXiv},
  author        = {Yue Zhang and Yafu Li and Leyang Cui and Deng Cai and Lemao Liu and Tingchen Fu and Xinting Huang and Enbo Zhao and Yu Zhang and Yulong Chen and Longyue Wang and Anh Tuan Luu and Wei Bi and Freda Shi and Shuming Shi},
  eprint        = {2309.01219},
  primaryclass  = {cs.CL},
  title         = {Siren's Song in the AI Ocean: A Survey on Hallucination in Large Language Models},
  year          = {2023}
}

@inproceedings{shi2024trusting,
  title     = {Trusting your evidence: Hallucinate less with context-aware decoding},
  author    = {Shi, Weijia and Han, Xiaochuang and Lewis, Mike and Tsvetkov, Yulia and Zettlemoyer, Luke and Yih, Wen-tau},
  booktitle = {Proceedings of the 2024 Conference of the North American Chapter of the Association for Computational Linguistics: Human Language Technologies (Volume 2: Short Papers)},
  pages     = {783--791},
  year      = {2024}
}

@article{kwiatkowski2019natural,
  title     = {Natural questions: a benchmark for question answering research},
  author    = {Kwiatkowski, Tom and Palomaki, Jennimaria and Redfield, Olivia and Collins, Michael and Parikh, Ankur and Alberti, Chris and Epstein, Danielle and Polosukhin, Illia and Devlin, Jacob and Lee, Kenton and others},
  journal   = {Transactions of the Association for Computational Linguistics},
  volume    = {7},
  pages     = {453--466},
  year      = {2019},
  publisher = {MIT Press One Rogers Street, Cambridge, MA 02142-1209, USA journals-info~…}
}

@article{li2022contrastive,
  title   = {Contrastive decoding: Open-ended text generation as optimization},
  author  = {Li, Xiang Lisa and Holtzman, Ari and Fried, Daniel and Liang, Percy and Eisner, Jason and Hashimoto, Tatsunori and Zettlemoyer, Luke and Lewis, Mike},
  journal = {arXiv preprint arXiv:2210.15097},
  year    = {2022}
}

@misc{jiang2023mistral7b,
  title         = {Mistral 7B},
  author        = {Albert Q. Jiang and Alexandre Sablayrolles and Arthur Mensch and Chris Bamford and Devendra Singh Chaplot and Diego de las Casas and Florian Bressand and Gianna Lengyel and Guillaume Lample and Lucile Saulnier and Lélio Renard Lavaud and Marie-Anne Lachaux and Pierre Stock and Teven Le Scao and Thibaut Lavril and Thomas Wang and Timothée Lacroix and William El Sayed},
  year          = {2023},
  eprint        = {2310.06825},
  archiveprefix = {arXiv},
  primaryclass  = {cs.CL},
  url           = {https://arxiv.org/abs/2310.06825}
}

@article{zhong2023mquake,
  title   = {Mquake: Assessing knowledge editing in language models via multi-hop questions},
  author  = {Zhong, Zexuan and Wu, Zhengxuan and Manning, Christopher D and Potts, Christopher and Chen, Danqi},
  journal = {arXiv preprint arXiv:2305.14795},
  year    = {2023}
}

@misc{guu2020realmretrievalaugmentedlanguagemodel,
  title         = {REALM: Retrieval-Augmented Language Model Pre-Training},
  author        = {Kelvin Guu and Kenton Lee and Zora Tung and Panupong Pasupat and Ming-Wei Chang},
  year          = {2020},
  eprint        = {2002.08909},
  archiveprefix = {arXiv},
  primaryclass  = {cs.CL},
  url           = {https://arxiv.org/abs/2002.08909}
}

@misc{izacard2021leveragingpassageretrievalgenerative,
  title         = {Leveraging Passage Retrieval with Generative Models for Open Domain Question Answering},
  author        = {Gautier Izacard and Edouard Grave},
  year          = {2021},
  eprint        = {2007.01282},
  archiveprefix = {arXiv},
  primaryclass  = {cs.CL},
  url           = {https://arxiv.org/abs/2007.01282}
}

@article{huang2025pip,
  title   = {PIP-KAG: Mitigating Knowledge Conflicts in Knowledge-Augmented Generation via Parametric Pruning},
  author  = {Huang, Pengcheng and Liu, Zhenghao and Yan, Yukun and Yi, Xiaoyuan and Chen, Hao and Liu, Zhiyuan and Sun, Maosong and Xiao, Tong and Yu, Ge and Xiong, Chenyan},
  journal = {arXiv preprint arXiv:2502.15543},
  year    = {2025}
}

@article{zhang2023alleviating,
  title   = {Alleviating hallucinations of large language models through induced hallucinations},
  author  = {Zhang, Yue and Cui, Leyang and Bi, Wei and Shi, Shuming},
  journal = {arXiv preprint arXiv:2312.15710},
  year    = {2023}
}

@article{zhang2024truthx,
  title   = {Truthx: Alleviating hallucinations by editing large language models in truthful space},
  author  = {Zhang, Shaolei and Yu, Tian and Feng, Yang},
  journal = {arXiv preprint arXiv:2402.17811},
  year    = {2024}
}

@misc{bai2022constitutionalaiharmlessnessai,
  title         = {Constitutional AI: Harmlessness from AI Feedback},
  author        = {Yuntao Bai and Saurav Kadavath and Sandipan Kundu and Amanda Askell and Jackson Kernion and Andy Jones and Anna Chen and Anna Goldie and Azalia Mirhoseini and Cameron McKinnon and Carol Chen and Catherine Olsson and Christopher Olah and Danny Hernandez and Dawn Drain and Deep Ganguli and Dustin Li and Eli Tran-Johnson and Ethan Perez and Jamie Kerr and Jared Mueller and Jeffrey Ladish and Joshua Landau and Kamal Ndousse and Kamile Lukosuite and Liane Lovitt and Michael Sellitto and Nelson Elhage and Nicholas Schiefer and Noemi Mercado and Nova DasSarma and Robert Lasenby and Robin Larson and Sam Ringer and Scott Johnston and Shauna Kravec and Sheer El Showk and Stanislav Fort and Tamera Lanham and Timothy Telleen-Lawton and Tom Conerly and Tom Henighan and Tristan Hume and Samuel R. Bowman and Zac Hatfield-Dodds and Ben Mann and Dario Amodei and Nicholas Joseph and Sam McCandlish and Tom Brown and Jared Kaplan},
  year          = {2022},
  eprint        = {2212.08073},
  archiveprefix = {arXiv},
  primaryclass  = {cs.CL},
  url           = {https://arxiv.org/abs/2212.08073}
}

@article{wang2024detoxifying,
  title   = {Detoxifying large language models via knowledge editing},
  author  = {Wang, Mengru and Zhang, Ningyu and Xu, Ziwen and Xi, Zekun and Deng, Shumin and Yao, Yunzhi and Zhang, Qishen and Yang, Linyi and Wang, Jindong and Chen, Huajun},
  journal = {arXiv preprint arXiv:2403.14472},
  year    = {2024}
}

@article{lin2021truthfulqa,
  title   = {Truthfulqa: Measuring how models mimic human falsehoods},
  author  = {Lin, Stephanie and Hilton, Jacob and Evans, Owain},
  journal = {arXiv preprint arXiv:2109.07958},
  year    = {2021}
}

@article{min2023factscore,
  title   = {Factscore: Fine-grained atomic evaluation of factual precision in long form text generation},
  author  = {Min, Sewon and Krishna, Kalpesh and Lyu, Xinxi and Lewis, Mike and Yih, Wen-tau and Koh, Pang Wei and Iyyer, Mohit and Zettlemoyer, Luke and Hajishirzi, Hannaneh},
  journal = {arXiv preprint arXiv:2305.14251},
  year    = {2023}
}

@article{wang2024deepedit,
  title   = {Deepedit: Knowledge editing as decoding with constraints},
  author  = {Wang, Yiwei and Chen, Muhao and Peng, Nanyun and Chang, Kai-Wei},
  journal = {arXiv preprint arXiv:2401.10471},
  year    = {2024}
}

@article{sinitsin2020editable,
  title   = {Editable neural networks},
  author  = {Sinitsin, Anton and Plokhotnyuk, Vsevolod and Pyrkin, Dmitriy and Popov, Sergei and Babenko, Artem},
  journal = {arXiv preprint arXiv:2004.00345},
  year    = {2020}
}

@article{yao2023editing,
  title   = {Editing large language models: Problems, methods, and opportunities},
  author  = {Yao, Yunzhi and Wang, Peng and Tian, Bozhong and Cheng, Siyuan and Li, Zhoubo and Deng, Shumin and Chen, Huajun and Zhang, Ningyu},
  journal = {arXiv preprint arXiv:2305.13172},
  year    = {2023}
}

@article{de2021editing,
  title   = {Editing factual knowledge in language models},
  author  = {De Cao, Nicola and Aziz, Wilker and Titov, Ivan},
  journal = {arXiv preprint arXiv:2104.08164},
  year    = {2021}
}

@article{cohen2024evaluating,
  title     = {Evaluating the ripple effects of knowledge editing in language models},
  author    = {Cohen, Roi and Biran, Eden and Yoran, Ori and Globerson, Amir and Geva, Mor},
  journal   = {Transactions of the Association for Computational Linguistics},
  volume    = {12},
  pages     = {283--298},
  year      = {2024},
  publisher = {MIT Press One Broadway, 12th Floor, Cambridge, Massachusetts 02142, USA~…}
}

@inproceedings{zhang2024sled,
  title     = {SLED: Self Logits Evolution Decoding for Improving Factuality in Large Language Models},
  author    = {Jianyi Zhang and Da-Cheng Juan and Cyrus Rashtchian and Chun-Sung Ferng and Heinrich Jiang and Yiran Chen},
  booktitle = {The Thirty-eighth Annual Conference on Neural Information Processing Systems (NeurIPS 2024},
  year      = {2024},
  url       = {https://arxiv.org/abs/2411.02433}
}

@article{gabriel2020artificial,
  title     = {Artificial intelligence, values, and alignment},
  author    = {Gabriel, Iason},
  journal   = {Minds and machines},
  volume    = {30},
  number    = {3},
  pages     = {411--437},
  year      = {2020},
  publisher = {Springer}
}

@article{ji2023ai,
  title   = {Ai alignment: A comprehensive survey},
  author  = {Ji, Jiaming and Qiu, Tianyi and Chen, Boyuan and Zhang, Borong and Lou, Hantao and Wang, Kaile and Duan, Yawen and He, Zhonghao and Zhou, Jiayi and Zhang, Zhaowei and others},
  journal = {arXiv preprint arXiv:2310.19852},
  year    = {2023}
}

@article{christiano2017deep,
  title   = {Deep reinforcement learning from human preferences},
  author  = {Christiano, Paul F and Leike, Jan and Brown, Tom and Martic, Miljan and Legg, Shane and Amodei, Dario},
  journal = {Advances in neural information processing systems},
  volume  = {30},
  year    = {2017}
}

@article{rafailov2023direct,
  title   = {Direct preference optimization: Your language model is secretly a reward model},
  author  = {Rafailov, Rafael and Sharma, Archit and Mitchell, Eric and Manning, Christopher D and Ermon, Stefano and Finn, Chelsea},
  journal = {Advances in neural information processing systems},
  volume  = {36},
  pages   = {53728--53741},
  year    = {2023}
}

@article{bai2022constitutional,
  title   = {Constitutional ai: Harmlessness from ai feedback},
  author  = {Bai, Yuntao and Kadavath, Saurav and Kundu, Sandipan and Askell, Amanda and Kernion, Jackson and Jones, Andy and Chen, Anna and Goldie, Anna and Mirhoseini, Azalia and McKinnon, Cameron and others},
  journal = {arXiv preprint arXiv:2212.08073},
  year    = {2022}
}

@article{askell2021general,
  title   = {A general language assistant as a laboratory for alignment},
  author  = {Askell, Amanda and Bai, Yuntao and Chen, Anna and Drain, Dawn and Ganguli, Deep and Henighan, Tom and Jones, Andy and Joseph, Nicholas and Mann, Ben and DasSarma, Nova and others},
  journal = {arXiv preprint arXiv:2112.00861},
  year    = {2021}
}

@article{casper2023open,
  title   = {Open problems and fundamental limitations of reinforcement learning from human feedback},
  author  = {Casper, Stephen and Davies, Xander and Shi, Claudia and Gilbert, Thomas Krendl and Scheurer, J{\'e}r{\'e}my and Rando, Javier and Freedman, Rachel and Korbak, Tomasz and Lindner, David and Freire, Pedro and others},
  journal = {arXiv preprint arXiv:2307.15217},
  year    = {2023}
}

@article{hendrycks2020aligning,
  title   = {Aligning ai with shared human values},
  author  = {Hendrycks, Dan and Burns, Collin and Basart, Steven and Critch, Andrew and Li, Jerry and Song, Dawn and Steinhardt, Jacob},
  journal = {arXiv preprint arXiv:2008.02275},
  year    = {2020}
}

@article{li2023self,
  title   = {Self-alignment with instruction backtranslation},
  author  = {Li, Xian and Yu, Ping and Zhou, Chunting and Schick, Timo and Levy, Omer and Zettlemoyer, Luke and Weston, Jason and Lewis, Mike},
  journal = {arXiv preprint arXiv:2308.06259},
  year    = {2023}
}

@inproceedings{park2023generative,
  title     = {Generative agents: Interactive simulacra of human behavior},
  author    = {Park, Joon Sung and O'Brien, Joseph and Cai, Carrie Jun and Morris, Meredith Ringel and Liang, Percy and Bernstein, Michael S},
  booktitle = {Proceedings of the 36th annual acm symposium on user interface software and technology},
  pages     = {1--22},
  year      = {2023}
}

@inproceedings{perez2023discovering,
  title     = {Discovering language model behaviors with model-written evaluations},
  author    = {Perez, Ethan and Ringer, Sam and Lukosiute, Kamile and Nguyen, Karina and Chen, Edwin and Heiner, Scott and Pettit, Craig and Olsson, Catherine and Kundu, Sandipan and Kadavath, Saurav and others},
  booktitle = {Findings of the association for computational linguistics: ACL 2023},
  pages     = {13387--13434},
  year      = {2023}
}

@article{madaan2023self,
  title   = {Self-refine: Iterative refinement with self-feedback},
  author  = {Madaan, Aman and Tandon, Niket and Gupta, Prakhar and Hallinan, Skyler and Gao, Luyu and Wiegreffe, Sarah and Alon, Uri and Dziri, Nouha and Prabhumoye, Shrimai and Yang, Yiming and others},
  journal = {Advances in Neural Information Processing Systems},
  volume  = {36},
  pages   = {46534--46594},
  year    = {2023}
}

@inproceedings{santurkar2023whose,
  title        = {Whose opinions do language models reflect?},
  author       = {Santurkar, Shibani and Durmus, Esin and Ladhak, Faisal and Lee, Cinoo and Liang, Percy and Hashimoto, Tatsunori},
  booktitle    = {International Conference on Machine Learning},
  pages        = {29971--30004},
  year         = {2023},
  organization = {PMLR}
}

@article{bang2024measuring,
  title   = {Measuring political bias in large language models: What is said and how it is said},
  author  = {Bang, Yejin and Chen, Delong and Lee, Nayeon and Fung, Pascale},
  journal = {arXiv preprint arXiv:2403.18932},
  year    = {2024}
}

@article{lin2022teaching,
  title   = {Teaching models to express their uncertainty in words},
  author  = {Lin, Stephanie and Hilton, Jacob and Evans, Owain},
  journal = {arXiv preprint arXiv:2205.14334},
  year    = {2022}
}

@article{fernando2023promptbreeder,
  title   = {Promptbreeder: Self-referential self-improvement via prompt evolution},
  author  = {Fernando, Chrisantha and Banarse, Dylan and Michalewski, Henryk and Osindero, Simon and Rockt{\"a}schel, Tim},
  journal = {arXiv preprint arXiv:2309.16797},
  year    = {2023}
}

@article{pryzant2023automatic,
  title   = {Automatic prompt optimization with" gradient descent" and beam search},
  author  = {Pryzant, Reid and Iter, Dan and Li, Jerry and Lee, Yin Tat and Zhu, Chenguang and Zeng, Michael},
  journal = {arXiv preprint arXiv:2305.03495},
  year    = {2023}
}

@article{kong2024aligning,
  title   = {Aligning large language models with representation editing: A control perspective},
  author  = {Kong, Lingkai and Wang, Haorui and Mu, Wenhao and Du, Yuanqi and Zhuang, Yuchen and Zhou, Yifei and Song, Yue and Zhang, Rongzhi and Wang, Kai and Zhang, Chao},
  journal = {Advances in Neural Information Processing Systems},
  volume  = {37},
  pages   = {37356--37384},
  year    = {2024}
}

@article{turner2023activation,
  title   = {Activation addition: Steering language models without optimization},
  author  = {Turner, Alexander Matt and Thiergart, Lisa and Udell, David and Leech, Gavin and Mini, Ulisse and MacDiarmid, Monte},
  journal = {CoRR},
  year    = {2023}
}

@article{liu2024decoding,
  title   = {Decoding-time realignment of language models},
  author  = {Liu, Tianlin and Guo, Shangmin and Bianco, Leonardo and Calandriello, Daniele and Berthet, Quentin and Llinares, Felipe and Hoffmann, Jessica and Dixon, Lucas and Valko, Michal and Blondel, Mathieu},
  journal = {arXiv preprint arXiv:2402.02992},
  year    = {2024}
}

@inproceedings{bender2021dangers,
  title     = {On the dangers of stochastic parrots: Can language models be too big?},
  author    = {Bender, Emily M and Gebru, Timnit and McMillan-Major, Angelina and Shmitchell, Shmargaret},
  booktitle = {Proceedings of the 2021 ACM conference on fairness, accountability, and transparency},
  pages     = {610--623},
  year      = {2021}
}

@article{zhang2024self,
  title   = {Self-exploring language models: Active preference elicitation for online alignment},
  author  = {Zhang, Shenao and Yu, Donghan and Sharma, Hiteshi and Zhong, Han and Liu, Zhihan and Yang, Ziyi and Wang, Shuohang and Hassan, Hany and Wang, Zhaoran},
  journal = {arXiv preprint arXiv:2405.19332},
  year    = {2024}
}

@article{liu2023trustworthy,
  title   = {Trustworthy llms: a survey and guideline for evaluating large language models' alignment},
  author  = {Liu, Yang and Yao, Yuanshun and Ton, Jean-Francois and Zhang, Xiaoying and Guo, Ruocheng and Cheng, Hao and Klochkov, Yegor and Taufiq, Muhammad Faaiz and Li, Hang},
  journal = {arXiv preprint arXiv:2308.05374},
  year    = {2023}
}

@incollection{pappas2025human,
  title     = {Human Value Alignment in AI},
  author    = {Pappas, Ilias O and Vassilakopoulou, Polyxeni},
  booktitle = {Handbook of Human-Centered Artificial Intelligence},
  pages     = {1--33},
  year      = {2025},
  publisher = {Springer}
}

@article{bhargava2023s,
  title   = {What's the magic word? a control theory of llm prompting},
  author  = {Bhargava, Aman and Witkowski, Cameron and Looi, Shi-Zhuo and Thomson, Matt},
  journal = {arXiv preprint arXiv:2310.04444},
  year    = {2023}
}

@article{bai2022training,
  title   = {Training a helpful and harmless assistant with rlhf},
  author  = {Bai, Yuntao and Kadavath, Saurav and Askell, Amanda and others},
  journal = {arXiv preprint arXiv:2204.05862},
  year    = {2022}
}

@article{alami2024alignment,
  title   = {Alignment with preference optimization is all you need for llm safety},
  author  = {Alami, Reda and Almansoori, Ali Khalifa and Alzubaidi, Ahmed and Seddik, Mohamed El Amine and Farooq, Mugariya and Hacid, Hakim},
  journal = {arXiv preprint arXiv:2409.07772},
  year    = {2024}
}

@inproceedings{schwenk2022okvqa,
  title        = {A-okvqa: A benchmark for visual question answering using world knowledge},
  author       = {Schwenk, Dustin and Khandelwal, Apoorv and Clark, Christopher and Marino, Kenneth and Mottaghi, Roozbeh},
  booktitle    = {Computer Vision--ECCV 2022: 17th European Conference, Tel Aviv, Israel, October 23--27, 2022, Proceedings, Part VIII},
  pages        = {146--162},
  year         = {2022},
  organization = {Springer}
}

@article{v-star,
  title   = {V*: Guided Visual Search as a Core Mechanism in Multimodal LLMs},
  author  = {Wu, Penghao and Xie, Saining},
  journal = {arXiv preprint arXiv:2312.14135},
  year    = {2023}
}

@article{li2023pope,
  title   = {Evaluating object hallucination in large vision-language models},
  author  = {Li, Yifan and Du, Yifan and Zhou, Kun and Wang, Jinpeng and Zhao, Wayne Xin and Wen, Ji-Rong},
  journal = {arXiv preprint arXiv:2305.10355},
  year    = {2023}
}

@inproceedings{textvqa,
  title     = {Towards vqa models that can read},
  author    = {Singh, Amanpreet and Natarajan, Vivek and Shah, Meet and Jiang, Yu and Chen, Xinlei and Batra, Dhruv and Parikh, Devi and Rohrbach, Marcus},
  booktitle = {Proceedings of the IEEE/CVF conference on computer vision and pattern recognition},
  pages     = {8317--8326},
  year      = {2019}
}

@article{qwen2025qwen25vl,
  title     = {Qwen2.5-VL: A Powerful Vision-Language Model for Seamless Computer Interaction},
  author    = {Qwen},
  journal   = {arXiv preprint arXiv:2409.12191},
  year      = {2025},
  publisher = {Alibaba Cloud}
}

@article{liu2023visual,
  title   = {Visual instruction tuning},
  author  = {Liu, Haotian and Li, Chunyuan and Wu, Qingyang and Lee, Yong Jae},
  journal = {arXiv preprint arXiv:2304.08485},
  year    = {2023}
}

@article{yuan2025autodrive,
  title={AutoDrive-R2: Incentivizing Reasoning and Self-Reflection Capacity for VLA Model in Autonomous Driving},
  author={Yuan, Zhenlong and Tang, Jing and Luo, Jinguo and Chen, Rui and Qian, Chengxuan and Sun, Lei and Chu, Xiangxiang and Cai, Yujun and Zhang, Dapeng and Li, Shuo},
  journal={arXiv preprint arXiv:2509.01944},
  year={2025}
}

@inproceedings{kirillov2023segment,
  title={Segment anything},
  author={Kirillov, Alexander and Mintun, Eric and Ravi, Nikhila and Mao, Hanzi and Rolland, Chloe and Gustafson, Laura and Xiao, Tete and Whitehead, Spencer and Berg, Alexander C and Lo, Wan-Yen and others},
  booktitle={Proceedings of the IEEE/CVF international conference on computer vision},
  pages={4015--4026},
  year={2023}
}

@inproceedings{redmon2016you,
  title={You only look once: Unified, real-time object detection},
  author={Redmon, Joseph and Divvala, Santosh and Girshick, Ross and Farhadi, Ali},
  booktitle={Proceedings of the IEEE conference on computer vision and pattern recognition},
  pages={779--788},
  year={2016}
}

@article{zhang2025mllms,
  title   = {Mllms know where to look: Training-free perception of small visual details with multimodal llms},
  author  = {Zhang, Jiarui and Khayatkhoei, Mahyar and Chhikara, Prateek and Ilievski, Filip},
  journal = {arXiv preprint arXiv:2502.17422},
  year    = {2025}
}

@article{chen2023combating,
  title={Combating misinformation in the age of llms: Opportunities and challenges},
  author={Chen, Canyu and Shu, Kai},
  journal={arXiv preprint arXiv:2311.05656},
  year={2023}
}

@inproceedings{mitchell2022memory,
  title={Memory-based model editing at scale},
  author={Mitchell, Eric and Lin, Charles and Bosselut, Antoine and Manning, Christopher D and Finn, Chelsea},
  booktitle={International Conference on Machine Learning},
  pages={15817--15831},
  year={2022},
  organization={PMLR}
}

@article{wang2023llm,
  title={An LLM-free Multi-dimensional Benchmark for MLLMs Hallucination Evaluation},
  author={Wang, Junyang and Wang, Yuhang and Xu, Guohai and Zhang, Jing and Gu, Yukai and Jia, Haitao and Yan, Ming and Zhang, Ji and Sang, Jitao},
  journal={arXiv preprint arXiv:2311.07397},
  year={2023}
}

@inproceedings{zhai-etal-2025-parameter,
    title = "Parameter-Aware Contrastive Knowledge Editing: Tracing and Rectifying based on Critical Transmission Paths",
    author = "Zhai, Songlin  and
      Meng, Yuan  and
      Zhang, Yuxin  and
      Qi, Guilin",
    editor = "Che, Wanxiang  and
      Nabende, Joyce  and
      Shutova, Ekaterina  and
      Pilehvar, Mohammad Taher",
    booktitle = "Proceedings of the 63rd Annual Meeting of the Association for Computational Linguistics (Volume 1: Long Papers)",
    month = jul,
    year = "2025",
    address = "Vienna, Austria",
    publisher = "Association for Computational Linguistics",
    url = "https://aclanthology.org/2025.acl-long.1367/",
    doi = "10.18653/v1/2025.acl-long.1367",
    pages = "28189--28200",
    ISBN = "979-8-89176-251-0"
}

@article{pan2024towards,
  title={Towards unified multimodal editing with enhanced knowledge collaboration},
  author={Pan, Kaihang and Fan, Zhaoyu and Li, Juncheng and Yu, Qifan and Fei, Hao and Tang, Siliang and Hong, Richang and Zhang, Hanwang and Sun, Qianru},
  journal={arXiv preprint arXiv:2409.19872},
  year={2024}
}
\bibliographystyle{IEEEtran}


\vspace{-1em}

\begin{IEEEbiography}[{\includegraphics[width=1in,height=1.1in,clip,keepaspectratio]{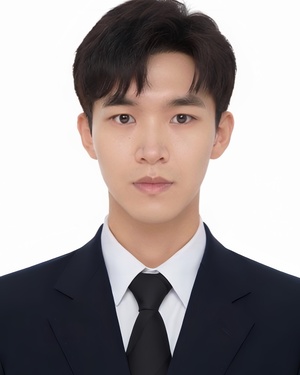}}]
{Baolong Bi} is a Ph.D. student in Computer Science at the Institute of Computing Technology, Chinese Academy of Sciences.  His research interests focus on the general area of trustworthy large language models.  He was a visiting researcher at the Language Technologies Institute, Carnegie Mellon University.  He has published 10+ top-tier papers, including ICLR, ICML, ACL, WWW, EMNLP, NAACL, etc, and serves as a reviewer for ICLR, ACL, COLM, NeurIPS, and CIKM.
\end{IEEEbiography}

\vspace{-3em}

\begin{IEEEbiography}[{\includegraphics[width=1in,height=1.1in,clip,keepaspectratio]{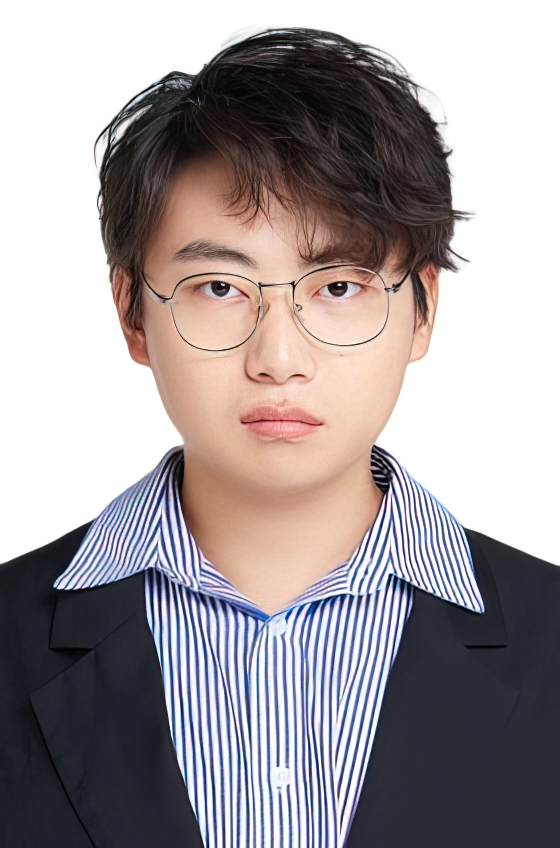}}]
{Yuyao Ge} is a Ph.D. student at the Institute of Computing Technology, Chinese Academy of Sciences, advised by Prof. Shenghua Liu. His research interests focus on the reasoning capabilities of large language models, particularly in vision-language models and complex logic problem solving. He has published papers in toptier conferences including ACL and EMNLP, and serves as a reviewer for AAAI and CIKM.
\end{IEEEbiography}

\vspace{-3em}

\begin{IEEEbiography}[{\includegraphics[width=1in,height=1.2in,clip,keepaspectratio]{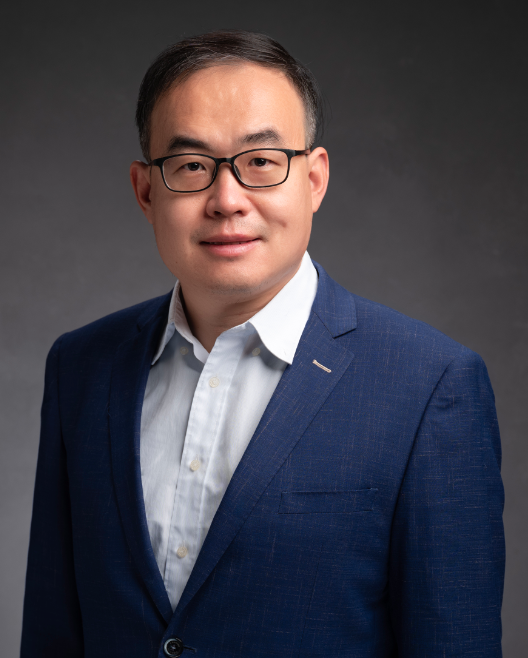}}]{Shenghua Liu} is a full Professor at the Institute of Computing
Technology, Chinese Academy of Sciences. He received his Ph.D. degree from the
Computer Science and Technology Department, Tsinghua University. He once
visited at the University of California, Los Angeles, hosted by Prof. Lei He, 
and as a scholar at Carnegie Mellon University, hosted by Prof. Christos Faloutsos. 
His current research focuses on big graph mining, large language models (LLMs), and scalable algorithm design, 
with recent interests in applying LLMs to graph analysis, and trustworthy foundation models. 
Two of about 60 high-quality publications have been recognized as
``best paper'' award and candidate respectively.
\end{IEEEbiography}

\vspace{-3em}

\begin{IEEEbiography}[{\includegraphics[width=1in,height=1.1in,clip,keepaspectratio]{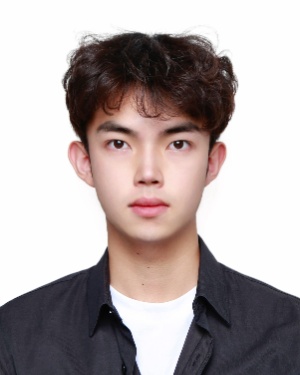}}]
{Yuchen He} is a senior undergraduate student majoring in Computer Science and Technology at Beijing Institute of Technology. He is currently a research intern at the Institute of Computing Technology, Chinese Academy of Sciences, advised by Assoc. Prof. Huaming Liao and Prof. Shenghua Liu. His research interests include trustworthy large language models and reinforcement learning with large language models. He has contributed to papers published in top-tier conferences including NeurIPS and SIGIR. 
\end{IEEEbiography}

\vspace{-3em}

\begin{IEEEbiography}[{\includegraphics[width=1in,height=1.1in,clip,keepaspectratio]{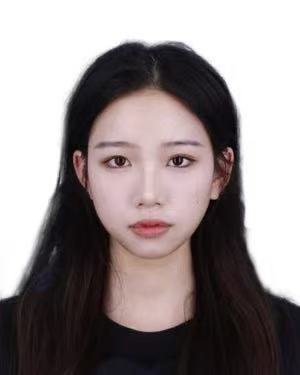}}]
{Siqian Tong} is a Ph.D. student in Artificial Intelligence in the University of Chinese Academy of Sciences, advised by Prof. Xuan Li.  Her research interests focus on the multimodal large language model.
\end{IEEEbiography}

\vspace{-3em}

\begin{IEEEbiography}[{\includegraphics[width=1in,height=1.1in,clip,keepaspectratio]{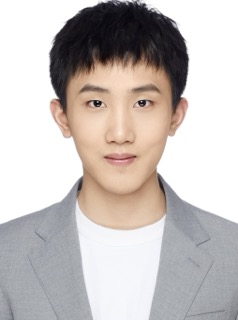}}]
{Lizhe Chen} is a Master's student in Computer Science at the Tsinghua University Shenzhen International Graduate School, advised by Prof. Shiguang Ni. His research interests focus on 3D data synthesis, combining large language models and computer graphics. He has published papers in top-tier conferences such as AAAI, IJCAI, and others, and has worked on research projects related to enhancing reasoning capabilities of large language models and 3D data synthesis. He also serves as a reviewer for conferences like AAAI, IJCAI, and others.
\end{IEEEbiography}

\vspace{-3em}

\begin{IEEEbiography}[{\includegraphics[width=1in,height=1.1in,clip,keepaspectratio]{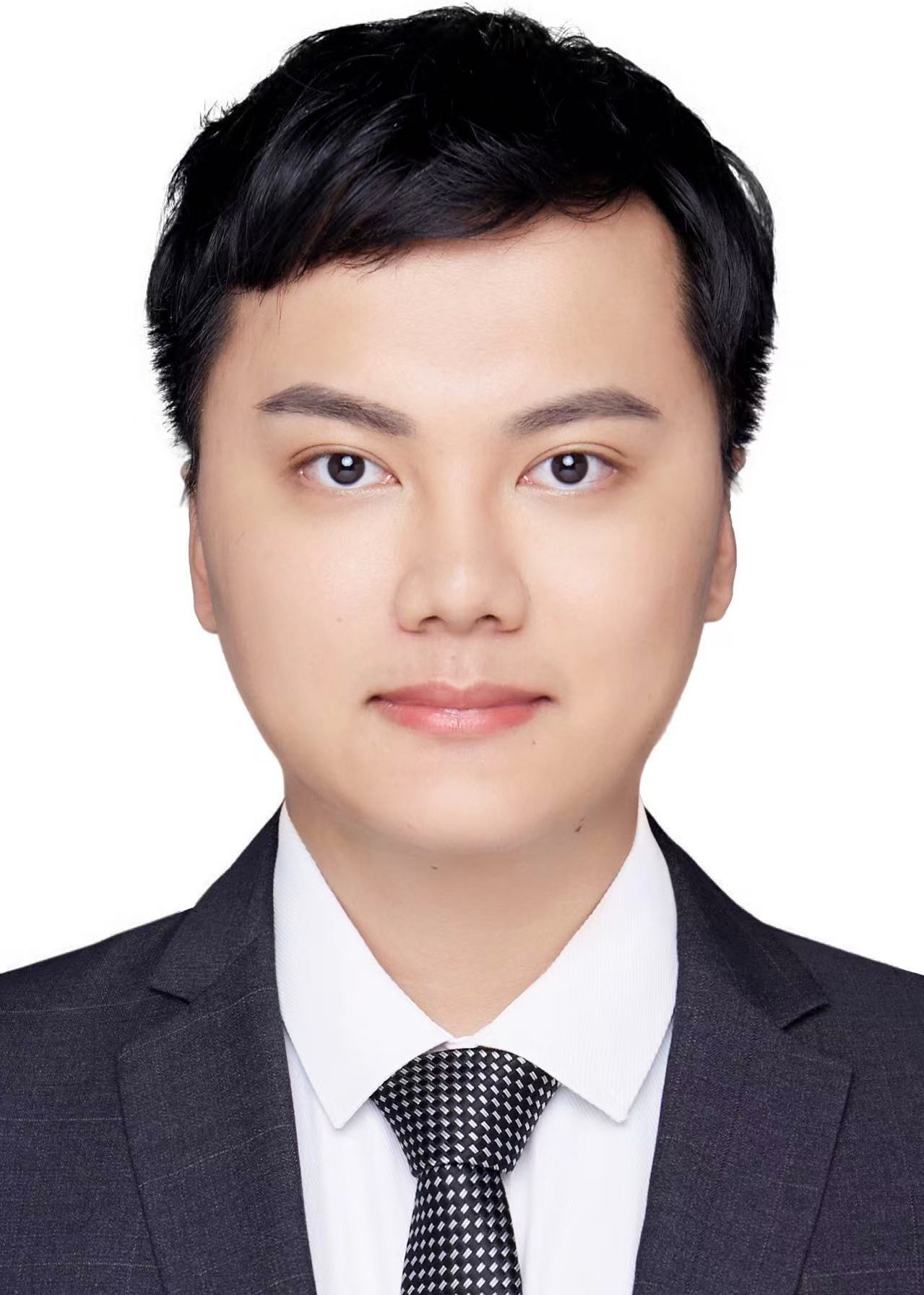}}]
{Lingrui Mei} is a Ph.D. student in Computer Science at the Institute of Computing Technology, Chinese Academy of Sciences, advised by Prof. Shenghua Liu. His research interests focus on large language models, including AI safety and agentic AI. He has published 10+ top-tier papers, including ICLR, ACL, EMNLP, etc, and serves as a reviewer for ICLR, ACL, COLM, NeurIPS, and EMNLP.
\end{IEEEbiography}

\vspace{-3em}

\begin{IEEEbiography}[{\includegraphics[width=1in,height=1.3in,clip,keepaspectratio]{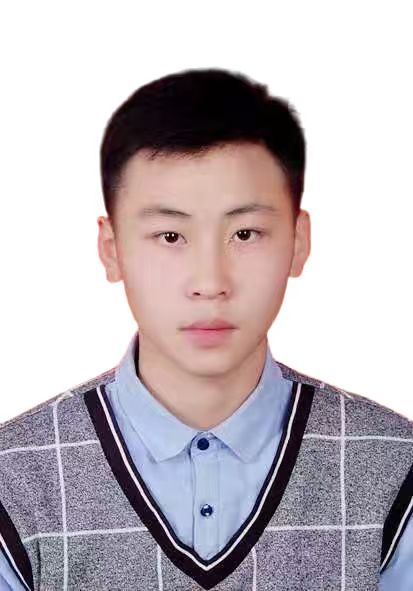}}]{Zehao Li}
received the B.Sc. degree in computer science and technology from Yunnan University, Kunming, China, in 2023. He is currently pursuing the Ph.D. degree with the Institute of Computing Technology, Chinese Academy of Sciences, Beijing, China. His main research interests include computer graphics and computer vision.
\end{IEEEbiography}

\vspace{-3em}

\begin{IEEEbiography}[{\includegraphics[width=1in,height=1.1in,clip,keepaspectratio]{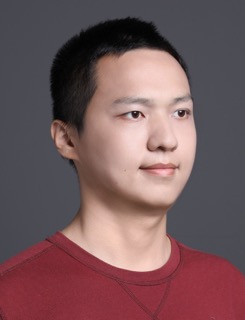}}]{Yiwei Wang}
is an Assistant Professor in the Computer Science Department at the University of California, Merced. 
His research focuses on natural language processing and graph machine learning, with recent work spanning large language model safety, multimodal reasoning, and retrieval-augmented generation. Dr. Wang has published extensively at top venues such as ACL, EMNLP, NeurIPS, ICML, ICLR, KDD, and AAAI, and his work has received recognitions including an EMNLP Best Paper Finalist and the SDSC Dissertation Research Fellowship. He serves as an Associate Editor for  \textit{IEEE TII}, \textit{Pattern Recognition}, and \textit{Neurocomputing}, and as a (Senior) Area Chair for major conferences including NeurIPS, ICML, ICLR, AAAI, ACL, EMNLP, COLM, IJCAI, and ICME. Additional information is available at \url{https://wangywust.github.io/}.
\end{IEEEbiography}

\vspace{-3em}

\begin{IEEEbiography}[{\includegraphics[width=1in,height=1.1in,clip,keepaspectratio]{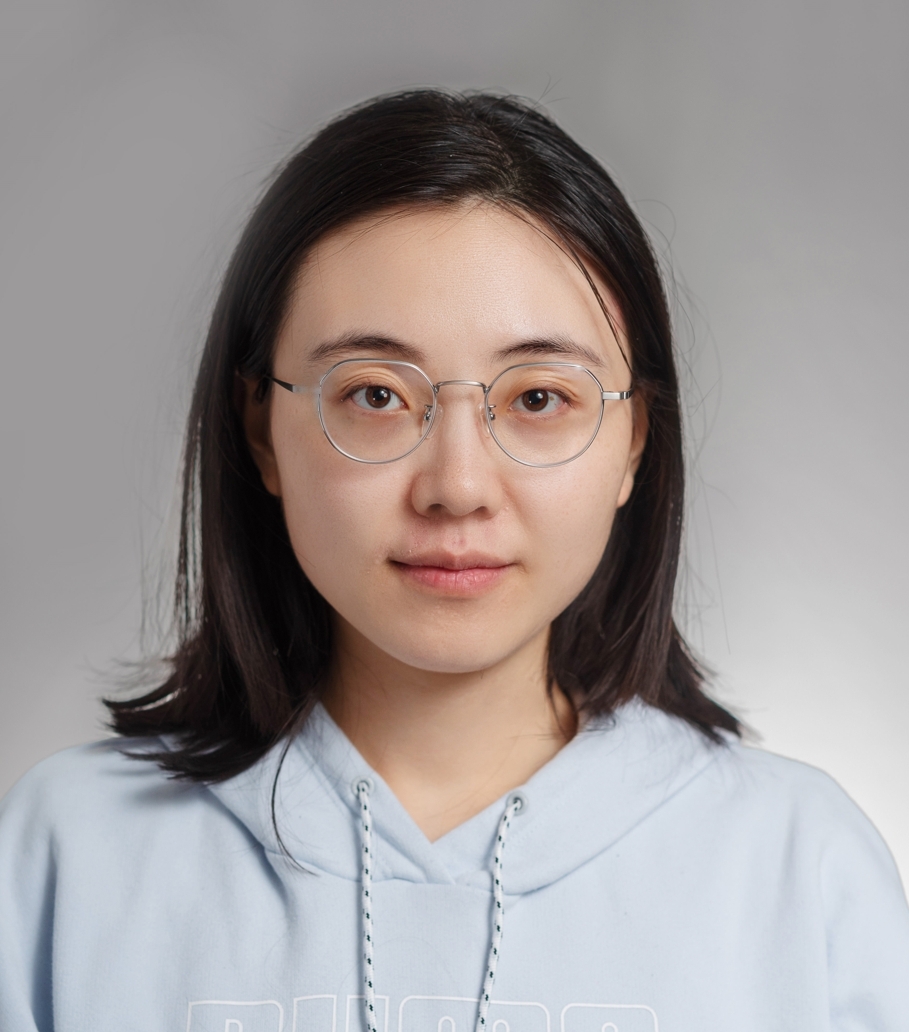}}]{Yujun Cai}
is currently a Lecturer at the School of Electrical Engineering and Computer Science, the University of Queensland (UQ), Australia. Before that she was a Research Scientist in Meta Inc USA. She obtained her Ph.D. from Nanyang Technological University in 2021. Her research focuses on multi-modal understanding, with publications in top-tier venues such as T-PAMI, CVPR, ICCV, NeurIPS, ECCV etc. She served as an Area Chair for conferences such as Neurips, CVPR, ICML, ICLR, ICCV, ACM MM, and as Associate Editor for journals such as Pattern Recognition and Neurocomputing.
\end{IEEEbiography}

\vspace{-3em}

\begin{IEEEbiography}[{\includegraphics[width=1in,height=1.2in,clip,keepaspectratio]{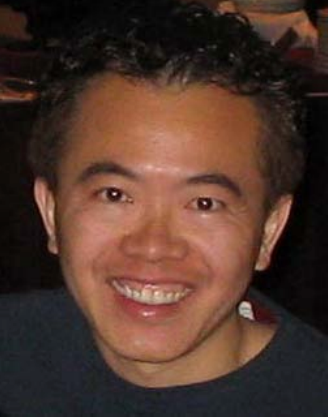}}]{Ming-Hsuan Yang}
(Fellow, IEEE) is affiliated with Google, UC Merced, and Yonsei University. Yang serves as a program co-chair of IEEE International Conference on Computer Vision (ICCV) in 2019, program co-chair of the Asian Conference on Computer Vision (ACCV) in 2014, and general co-chair of ACCV 2016. Yang served as an associate editor of the IEEE Transactions on Pattern Analysis and Machine Intelligence and is an associate editor of the International Journal of Computer Vision, Image and Vision Computing and Journal of Artificial Intelligence Research. He received the NSF CAREER award and Google Faculty Award. He is a Fellow of the IEEE, ACM, and AAAI.
\end{IEEEbiography}

\vspace{-3em}

\begin{IEEEbiography}[{\includegraphics[width=1in,height=1.2in,clip,keepaspectratio]{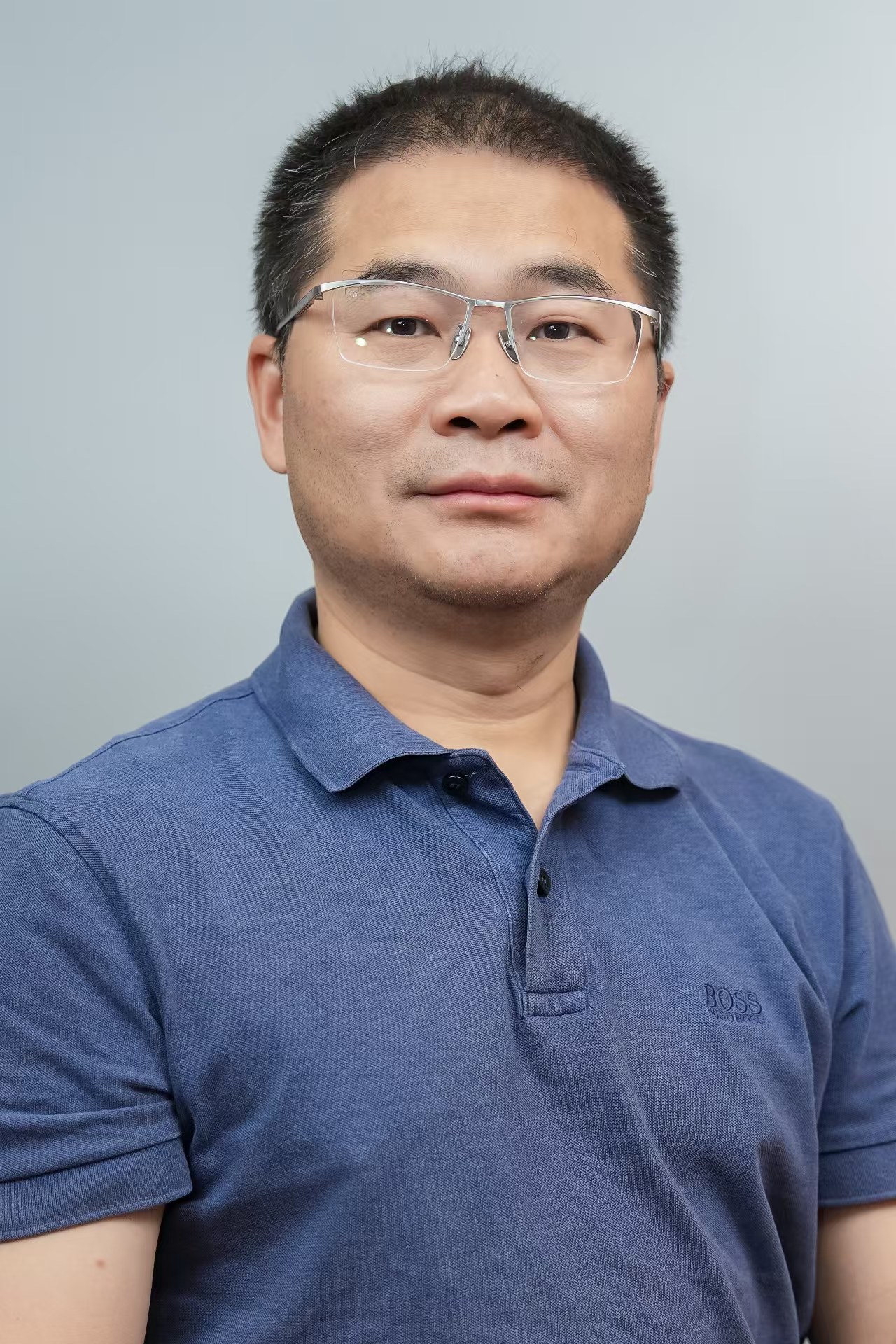}}]{Xueqi Cheng}
(Fellow, IEEE) is a professor with the Institute of Computing Technology, Chinese Academy of Sciences (ICT-CAS) and the University of Chinese Academy of Sciences, and the director of the CAS Key Laboratory of Network Data Science and Technology. His main research interests include network science, web search and data mining, Big Data processing and distributed computing architecture. He has won the Best Paper Award in CIKM (2011), the Best Student Paper Award in SIGIR (2012), and the Best Paper Award Runner up of CIKM (2017).
\end{IEEEbiography}

\vfill

\end{document}